# Network representation learning: A macro and micro view

Xueyi Liu [a], Jie Tang [a,b,*]

[a] Department of Computer Science and Technology, Tsinghua University, Beijing, 100084, China
[b] Tsinghua University, Tsinghua National Laboratory for Information Science and Technology (TNList), Tsinghua-Bosch Joint ML Center, Beijing, 100084, China

Abstract

Graph is a universe data structure that is widely used to organize data in real-world. Various real-word networks like the transportation network, social and academic network can be represented by graphs. Recent years have witnessed the quick development on representing vertices in the network into a low-dimensional vector space, referred to as network representation learning. Representation learning can facilitate the design of new algorithms on the graph data. In this survey, we conduct a comprehensive review of current literature on network representation learning. Existing algorithms can be categorized into three groups: shallow embedding models, heterogeneous network embedding models, graph neural network based models. We review state-of-the-art algorithms for each category and discuss the essential differences between these algorithms. One advantage of the survey is that we systematically study the underlying theoretical foundations underlying the different categories of algorithms, which offers deep insights for better understanding the development of the network representation learning field.

## 1. Introduction

Graph is a highly expressive data structure, based on which various networks exist in the real-world, like the social networks (Aggarwal, 2011; Myers et al., 2014), citation networks (Sen et al., 2008), biological networks (Marinka et al., 2019), chemistry networks (Martins et al., 2012), traffic networks, and others. Mining information from real-world networks plays a crucial role in many emerging applications. For example, in social networks, classifying people into social communities according to their profile and social connections is useful for many related task, like social recommendation, target advertising, user search (Zhang et al., 2018), etc. In communication networks, detecting community structures can help understand information diffusion. In biological networks, predicting the role of protein can help us reveal the mysteries of life; predicting molecular drugability can promote new drug development. In chemistry networks, predicting the function of molecules can help with the synthesis of new compound and new material. The way in which networks are generally represented cannot supply effective analysis. For example, the only structural information we can get from an adjacency matrix about one node is just its neighbours and the weight of the edges between them. It is not informative enough with respect to the neighbourhood structure and its role in the graph, and also of high space complexity (i.e., $O(N)$ for one node, where $N$ is the number of nodes in the network). It is also hard to design an efficient algorithm based just on the adjacency matrix. Taking community detection as an example, most existing algorithms will involve calculating the spectral decomposition of a matrix (Fragkiskos and Vazirgiannis, 2013), whose time complexity is always at least quadratic with respect to the number of vertices. Existing graph analytical methods, like distributed graph data processing framework (e.g., GraphX (Gonzalez et al., 2014), and GraphLab (LowYet al., 2012)) suffer from high computational cost and high space complexity. This complexity makes the algorithms hard to be applied to large-scale networks with millions of vertices.

Recent years have seen the rapid development of network representation learning algorithms. Their purpose is to learn latent, informative and low-dimensional representations for network vertices, which can preserve the network structure, vertex features, labels and other auxiliary information (Cai et al., 2018; Zhang et al., 2018), as Fig. 1 illustrates. The vertex representations can help design efficient algorithms since various vector based machine learning algorithms can thus be easily applied to vertex representation vectors.

Such works date back to the early 2000s (Zhang et al., 2018), when the proposed algorithms were part of dimensionality-reduction techniques (e.g., Isomap (B Tenenbaum et al., 2000), LLE (Roweis and Lawrence, 2000), Eigenmap (Belkin and Niyogi, 2002), and MFA (Yan et al., 2007)). These algorithms firstly calculate the affinity graph (e.g., $k$-nearest-neighbour graph) for the input of high-dimensional data. Then, the affinity graph is embedded into a lower dimensional space. However, the time complexity of those methods is too high to scale to large networks. Later on, there is an emerging number of works (Bryan et al., 2014; Zhang et al., 2016a; Cao et al., 2016) focusing on developing efficient and effective embedding method to assign each node a low dimensional representation vector that is aware of structural information, vertex content and other information. Many efficient machine learning models can be designed for downstream tasks based on the learned vertex representations, like node classification (Zhu et al., 2007; BhagatGraham and Muthukrishnan, 2011), link prediction (Linyuan and Zhou, 2011; Gao et al., 2011; Liben-Nowell and Kleinberg, 2003), recommendation (Zhang et al., 2017; Xie et al., 2016), similarity search (Liu et al., 2018), visualization (Tang et al., 2016), clustering (Fragkiskos and Vazirgiannis, 2013), and knowledge graph search (Lin et al., 2015). Fig. 2 shows a brief summary of the development history of graph embedding models.

* Corresponding author. Department of Computer Science and Technology, Tsinghua University, Tsinghua-Bosch Joint ML Center, Beijing, China.
*E-mail address:* jietang@tsinghua.edu.cn (J. Tang).

In this survey, we provide a comprehensive up-to-date review of network representation learning algorithms, aiming to give readers a macro, covering the some common basic insights under different kinds of embedding algorithms and the relationship between them, as well as a micro, lacking no details of different algorithms and also theories behind them, view on previous effort and achievements in this area. We group existing graph embedding methods into three major categories based on the development dependencies among those algorithms, from *shallow embedding models*, whose objects are basic homogeneous graphs (Def. 1) with only one type of nodes and edges,[1] to *heterogeneous embedding models*, most of whose basic ideas are inherited from shallow embedding models designed for homogeneous graphs with the range of graph objects expanded to heterogeneous graphs with more than one types of nodes or edges and also often node or edge features, then further to *graph neural network based models*, many of whose insights are able to be found in shallow embedding models and heterogeneous embedding models, like the inductive learning and neighbourhood aggregation (Cen et al., 2019), spectral propagation (Donnat et al., 2017; Zhang et al., 2019b), and so on. Though it is hard to say the ideas of which methods are inspired by whose thoughts, the similarity and connections between them can help us understand them better and also always offer some interesting rethinking of the common field they belong to, which are also what the survey focuses on beyond reviewing existing graph embedding techniques.

Table 1 lists some typical graph embedding models and some of their related information, which can help readers get a fast glimpse of existing graph embedding models, their inner mechanisms and underlying relations. *Shallow embedding models* can be roughly grouped into two main categories, shallow neural embedding models and matrix factorization based models. Shallow neural embedding models (S–N) are characterized by embedding look-up tables, which are updated to preserve various proximities lying in the graph. Typical models are DeepWalk (Bryan et al., 2014), node2vec (Grover and Leskovec, 2016), LINE (Tang et al., 2015b) and so on. Matrix factorization (S-MF) based models aim to factorize matrices related with graph structure and other side information to get high-quality node representation vectors. Based on shallow embedding models designed for homogeneous networks, embedding techniques (e.g. PTE (Tang et al., 2015a), metapath2vec (Dong et al., 2017), GATNE (Cen et al., 2019)), are designed for heterogeneous networks and we refer these models to *heterogeneous (SH) embedding models*. Different from shallow embedding models, *graph neural networks (GNNs)* are kind of techniques characterized by deep architectures to extract meaningful structural information into node representation vectors. In addition to the discussion of the above-mentioned types of models, we also focus on their inner connections, advantages and disadvantages, optimization methods and some related theoretical foundations.

Finally, we summarize some existing challenges and propose possible development directions that can help with further design. We organize the survey as follows. In Section 2, we first summarize some useful definitions which can help readers understand the basic concepts, and then propose our taxonomy for the existing embedding algorithms. Then, in Section 4, 5, and 6, we review typical embedding methods falling into those three categories. In Section 7 and 8 we discuss some relationships within those algorithms of different categorizes and related optimization methods. We then go further to discuss some problems and challenges of existing graph embedding models in Section 9. At last, we discuss some further development directions for network representation learning in Section 10.

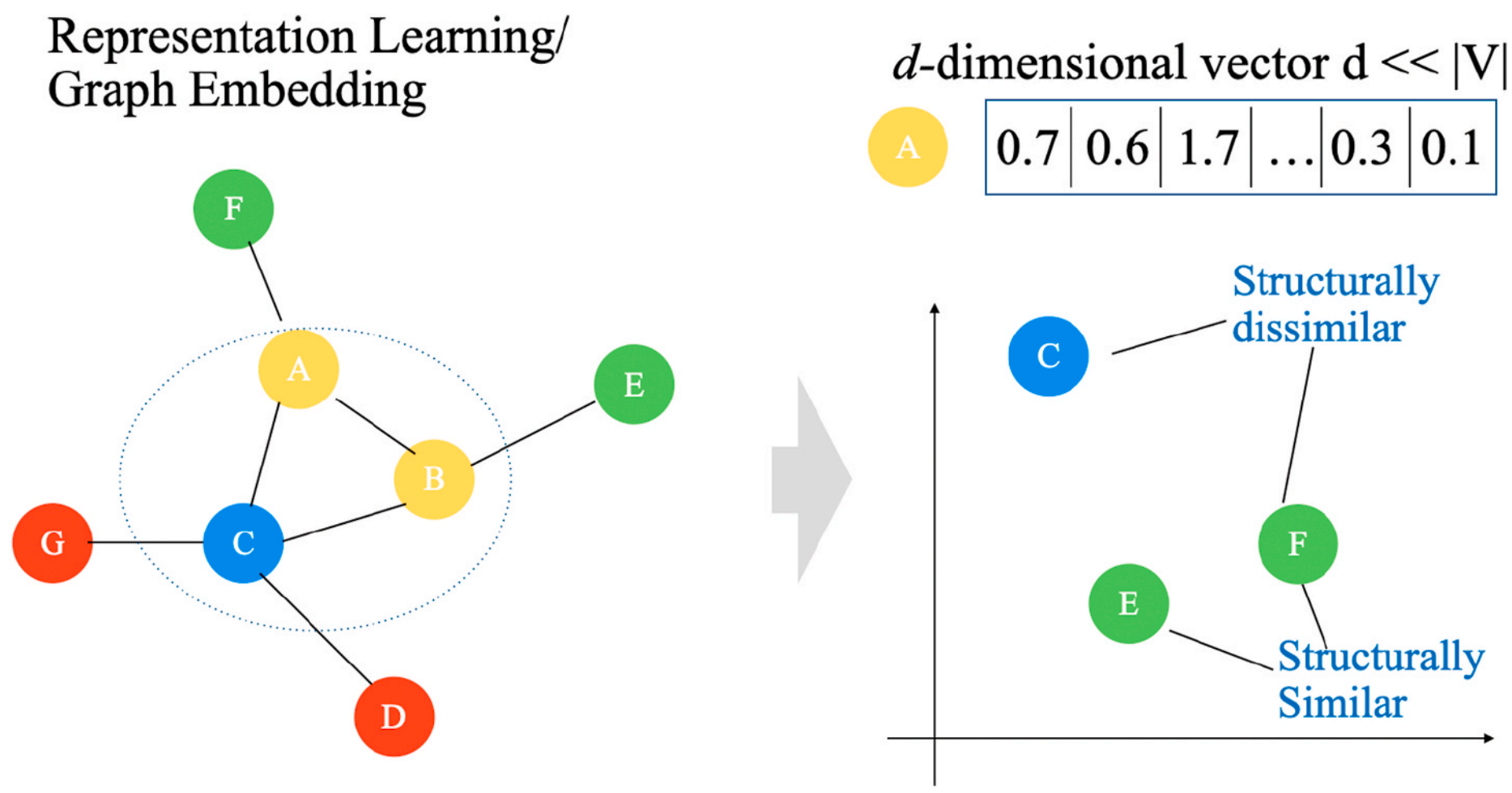

**Fig. 1.** A toy example for network embedding task. Vertices in the network lying in the left part are embedded into $d$-dimensional vector space, where $d$ is much smaller than the total number of nodes $|V|$ in the network. Vertices with the same color are structurally similar to each other. Basic structural information should be kept in the embedding space (e.g., Structurally similar vertices E and F are embedded closer to each other than structurally dissimilar vertices C and F).

[1] The word "homogeneous" is omitted in the category name "shallow embedding models" for brevity.

## 2. Preliminaries

We summarize related definitions as follows to help readers understand the algorithms discussed in the following parts.

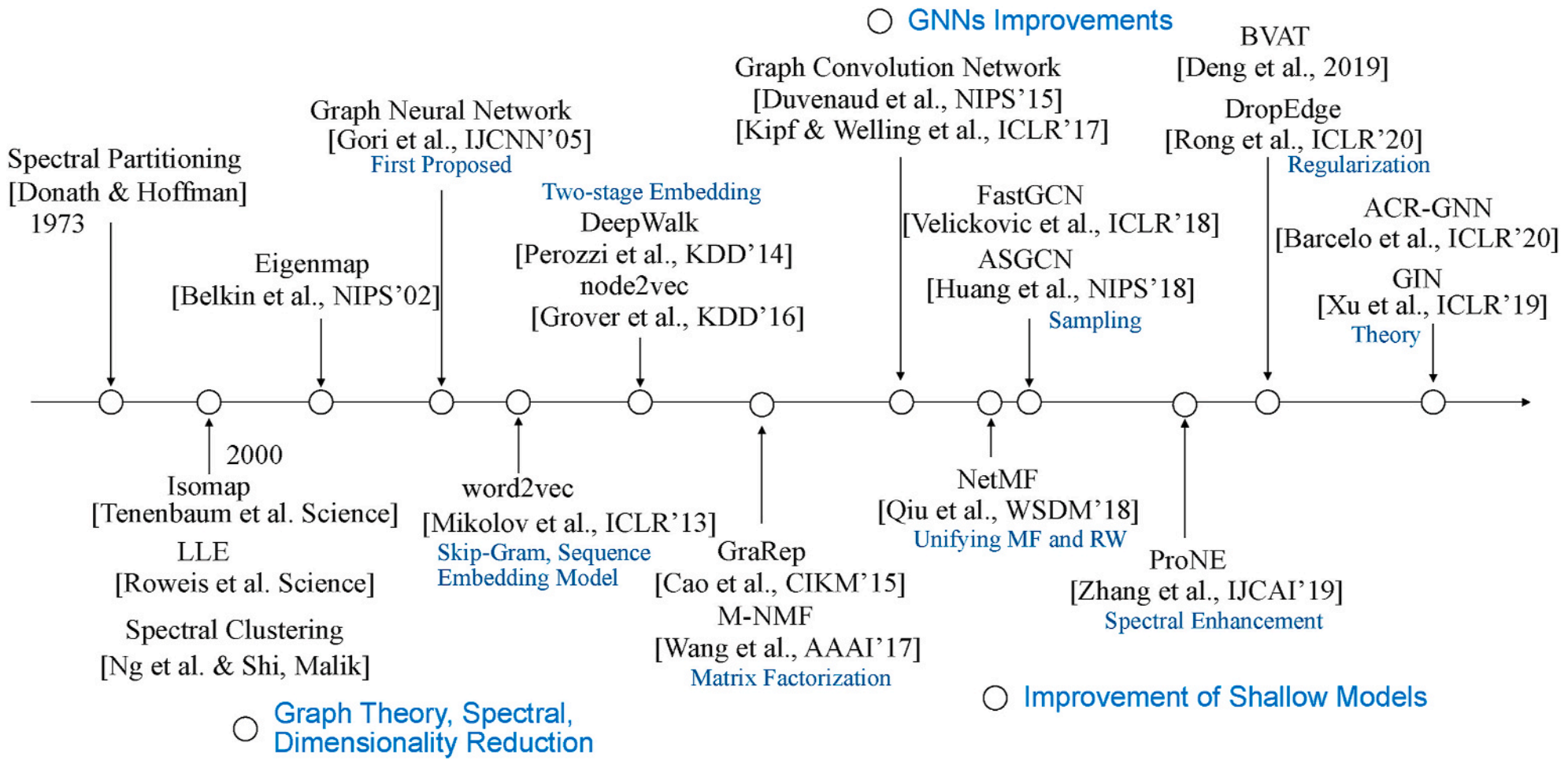

**Fig. 2.** A brief summary of the development of network embedding techniques. Left Channel: Shallow (Heterogeneous) Neural Embedding Models; Mid Channel: Matrix Factorization Based Models; Right Channel: Graph Neural Network Based Models.

First we introduce the definition of a graph, which is the basic data structure of real-world networks:

**Definition 1**. (Graph). A graph can be denoted as $G = (\mathscr{V}, \mathscr{E})$, where $\mathscr{V}$ is the set of vertices and $\mathscr{E}$ is the set of edges in the graph. When associated with the node type mapping function $\Phi: \mathscr{V} \rightarrow \mathscr{O}$ mapping each node to its specific node type and an edge mapping function $\Psi: \mathscr{E} \rightarrow \mathscr{R}$ mapping each edge to its corresponding edge type, a graph G can be divided into two categories: homogeneous graph and heterogeneous graph. A homogeneous graph is a graph G with only one node type and one edge type (i.e., $|\mathscr{O}| = 1$ and $|\mathscr{R}| = 1$). A graph is a heterogeneous graph when $|\mathscr{O}| + |\mathscr{R}| > 2$.

Graphs are basic data structure for many kinds of real-world networks, like transportation network (Ribeiro et al., 2017), social networks, academic networks (Scott et al., 2016; Yang and Leskovec, 2015), and so on. They can be modeled by homogeneous graphs or heterogeneous graphs, based on the knowledge we have on nodes and edges × in those networks. In the survey, we use graph embedding and network representation learning alternatively, both of which are high-frequency terms appeared in the literature (Zhang et al., 2019b; Cen et al., 2019; Bryan et al., 2014; Grover and Leskovec, 2016; Yang et al., 2020b) and both denote the process of generating representative vectors of a finite dimension for nodes in a graph or a network. When we use the term graph embedding, we focus mainly on the basic graph models, where we simply care about nodes and edges in the graph, and when we use network representation learning, our focus is more on networks in real-world.

Since there is a large number of embedding algorithms based on modeling vertex proximities, we briefly summarize the proposed vertex similarities as follows (Zhang et al., 2018):

**Definition 2**. (Vertex Proximities). Various vertex proximities can exist in real-world networks, like first-order proximity, second-order proximity and higher-order proximities. The first-order proximity can measure the direct connectivity between two nodes, which is usually defined as the weight of the edge between them. The second-order proximity between two vertices can be defined as the distance between the distributions of their neighbourhood (Wang et al., 2016). Higher-order proximities between two vertices v and u can be defined as the k-step transition probability from vertex v to vertex u (Zhang et al., 2018).

**Definition 3**. (Structural Similarity). Structural similarity (Ribeiro et al., 2017; Henderson et al., 2012; Donnat et al., 2017; Lorrain and White, 1977; Pizarro, 2007) refers to the similarity of the structural roles of two vertices in their respective communities, although they may not connect with each other.

**Definition 4**. (Intra-community Similarity). The intra-community similarity originates from the community structure of the graph and denotes the similarity between two vertices that are in the same community. Many real-life networks (e.g., social networks, citation networks) have community structure, where vertex-vertex connections in a community are dense, but sparse for nodes between two communities.

Since graph Laplacian matrices are based for understanding embedding algorithms based on graph spectral, or adopt the graph spectral way, which is also a crucial development direction for embedding algorithms, we briefly introduce them as follows:

**Definition 5**. (Graph Laplacian). Following notions in (Hoang and Maehara, 2019), $\boldsymbol{L} = \boldsymbol{D} - \boldsymbol{A}$, where $\boldsymbol{A}$ is the adjacency matrix, D is the corresponding degree matrix, is the combinational graph laplacian, $\mathscr{L} = \boldsymbol{I} - \boldsymbol{D}^{-\frac{1}{2}}\boldsymbol{A}\boldsymbol{D}^{-\frac{1}{2}}$ is the normalized graph Laplacian, $\boldsymbol{L}_{rw} = \boldsymbol{I} - \boldsymbol{D}^{-1}\boldsymbol{A}$ is the random walk graph Laplacian. Meanwhile, let $\widetilde{\boldsymbol{A}} = \boldsymbol{A} + \sigma \boldsymbol{I}$ denotes the augmented adjacency matrix, then, $\widetilde{\boldsymbol{L}}, \widetilde{\mathscr{L}}, \widetilde{\boldsymbol{L}}_{rw}$ are the augmented graph Laplacian, augmented normalized graph Laplacian, augmented random walk graph Laplacian respectively.

## 3. Overview of graph embedding techniques

In this section, we will give graph embedding techniques of each category a brief introduction to help readers get a better understanding of the overall architecture of this paper. Fig. 3 shows a panoramic view of existing embedding models and their connections.

**Shallow Embedding Models.** These models can be divided into two main streams: shallow neural embedding models and matrix factorization based models. Though there are some differences between those two embedding genres, it has been shown that some shallow embedding based models, especially those adopt random walk to sample vertex sequences and perform skip-gram model to get vertex embeddings, have close connections with matrix factorization models: they are actually implicitly factorizing their equivalent matrices (Qiu et al., 2018a), to be specific.

Besides, matrices being factorized by shallow embedding models also have close relationship with graph spectral theories. Apart from models like GraphWave (Donnat et al., 2017) which are based on graph spectral directly (see Sec. 4.4), other models like DeepWalk (Bryan et al., 2014), node2vec (Grover and Leskovec, 2016), LINE (Tang et al., 2015b) can also be proved to have close relationship with graph spectral by proving that their equivalent matrices are filter matrices (Qiu et al., 2018a).

Then, the explicit combination of traditional shallow embedding

**Table 1**
An overview of network representation learning algorithms (selected). Symbols in some formulas can refer to Def. 5. For others, “A” ~ w/o vertex attributes;“L” ~ w/o vertex labels; “Heter.” ~ heterogeneous networks. Abbreviations used: “F-O”, “S-O”, “H-O”, “I-C”, “ST” refer to First-Order, Second-Order, High-Order, Intra-Community and Structural similarities; “SN”, “SHN”, “MF”, “SS” refer to Shallow Neural Embedding models, Shallow Heterogeneous Network Embedding Model, Matrix Factorization Based models and Shallow Spectral models; “O-S” denotes optimization strategies, in which “PS”, “NS” refer to positive sampling and negative sampling; “(r)-(t)SVD” refers to (randomized)-(truncated) singular value decomposition; “SGNS” refers to “Skip-Gram with Negative Sampling”; “Iter-Update” refers to.

| Model | Type | Neural | Heter. | A/L | O-S | Proximity | Matrix | Filter |
|---|---|---|---|---|---|---|---|---|
| DeepWalk (Bryan et al., 2014) | S-N | ✓ | × | × | SGNS,G | H-O | Table 2 DeepWalk | $h(\lambda) = \frac{1}{T}\sum_{r=1}^{T}\lambda^r$ |
| node2vec (Grover and Leskovec, 2016) | | ✓ | × | × | SGNS,G | | Table 2 node2vec | - |
| Diff2vec (Rozemberczki and Sarkar, 2020) | | ✓ | × | × | PS | | - | - |
| Walklets (Bryan et al., 2016) | | ✓ | × | × | SGNS,G | | - | - |
| Rol2Vec (AhmedRyan et al., 2018) | | ✓ | × | A | SGNS,G | I-C | - | - |
| LINE (Tang et al., 2015b) | | ✓ | × | × | PS,NS,D,G | F-O,S-O | Table 2 LINE | $h(\lambda) = 1 - \lambda$ |
| pRBM [138] | | ✓ | × | A | PS,G | F-O | - | - |
| UPP-SNE (Zhang et al., 2016c) | | ✓ | × | A | SGNS,G | H-O | - | - |
| DDRW (Li et al., 2016) | | ✓ | × | L | SGNS,SVM D,G | H-O | Table 2 DeepWalk | $h(\lambda) = \frac{1}{T}\sum_{r=1}^{T}\lambda^r$ |
| TLINE (Wu et al., 2019) | | ✓ | × | L | PS,NS,SVM D,G | F-O S-O | Table 2 LINE | $h(\lambda) = 1 - \lambda$ |
| GraphGAN (Tang et al., 2015b) | | ✓ | × | × | G,D | F-O | - | - |
| struct2vec (Natarajan and Inderjit, 2014) | | ✓ | × | × | SGNS,D | ST | - | - |
| PTE (Tang et al., 2015a) | SH | ✓ | ✓ | × | PS,NS,G | S-O | Eq. 13 | - |
| metapath2vec (Dong et al., 2017) | | ✓ | ✓ | × | SGNS | H-O | - | - |
| HIN2vec (AhmedRyan et al., 2018) | | ✓ | ✓ | × | SGNS | H-O | - | - |
| GATNE (Cen et al., 2019) | | ✓ | ✓ | A | SGNS | H-O | - | - |
| HERec (Shi et al., 2019) | | ✓ | ✓ | L | SGNS MF | H-O | user rating matrix $\boldsymbol{R}$ | - |
| HueRec (Wang et al., 2019b) | | ✓ | ✓ | L | SGNS | H-O | - | - |
| HeGAN (HuYuan and Shi, 2019) | | ✓ | ✓ | × | G,D | F-O | - | - |
| M-NMF (Wang et al., 2017b) | S-MF | × | × | × | Iter-Update | F-O,S-O, I-C | $\boldsymbol{S} = \boldsymbol{S}^{(1)} + \eta \boldsymbol{S}^{(2)}, \boldsymbol{H}$ | - |
| NetMF (Qiu et al., 2018a) | | × | × | × | tSVD | H-O | Table 2 DeepWalk | $h(\lambda) = \frac{1}{T}\sum_{r=1}^{T}\lambda^r$ |
| ProNE (Zhang et al., 2019b) | | × | × | × | r-tSVD | F-O | Eq. 5 | - |
| GraRep (Cao et al., 2015) | | × | × | × | tSVD | H-O | Eq. 6 | - |
| HOPE (Ou et al., 2016) | | × | × | × | JDGSVD [Hochstenbach, 2009] | H-O | General | - |
| TADW (Cheng et al., 2015) | | × | × | A | I-MF | H-O(without homophily) | $k$-step transition matrix $\boldsymbol{M}$ | - |
| HSCA (Zhang et al., 2016b) | | × | × | A | Iter-Update | H-O | $k$-step transition matrix $\boldsymbol{M}$ | - |
| ProNE (Zhang et al., 2019b) | S-SS | × | × | × | | H-O | $\boldsymbol{I}_N - \boldsymbol{U}g(\Lambda)\boldsymbol{U}^{-1}$ | $g(\lambda) = e^{-\frac{1}{2}[(\lambda-\mu)^2-1]\theta}$ |
| GraphZoom (Deng et al., 2020) | | × | × | A | Spectral Propagation | H-O | $\left(\widetilde{\boldsymbol{D}}^{-\frac{1}{2}}\widetilde{\boldsymbol{A}}\widetilde{\boldsymbol{D}}^{-\frac{1}{2}}\right)^k$ | $h(\lambda) = (1-\lambda)^k$ |
| GraphWave (Donnat et al., 2017) | | × | × | × | | ST | - | $g_s(\lambda) = e^{-\lambda s}$ |
| GCN (Kipf and Welling, 2017) | GNN | ✓ | × | A,L | SGD | H-O | $\widetilde{\boldsymbol{D}}^{-\frac{1}{2}}\widetilde{\boldsymbol{A}}\widetilde{\boldsymbol{D}}^{-\frac{1}{2}}$ | $h(\lambda) = 1 - \lambda$ |
| GraphSAGE (Hamilton et al., 2017) | | ✓ | × | A,L | SGD | H-O | Depend on A-M | - |
| FastGCN (Chen et al., 2018) | | ✓ | × | A,L | SGD | H-O | $\widetilde{\boldsymbol{A}}Q\left(Q_{ii} = \frac{1}{q_{v_i}^{(l)}}\right)$ | $h(\lambda) = 1 - \lambda$ |
| ASGCN (Huang et al., 2018) | | ✓ | × | A,L | SGD | H-O | $\widetilde{\boldsymbol{A}}\boldsymbol{Q}\left(Q_{ii} = \frac{q_i}{q*_{v_i}}\right)$ | $h(\lambda) = 1 - \lambda$ |
| GAT (Veličković et al., 2018) | | ✓ | × | A,L | SGD | H-O | $\boldsymbol{P}\widetilde{\boldsymbol{A}} + \widetilde{\boldsymbol{A}}\boldsymbol{Q}$[5] | - |
| GIN (Vaswani et al., 2017) | | ✓ | × | A,L | SGD | H-O | $\widetilde{\boldsymbol{A}} = \boldsymbol{A} + \boldsymbol{I}_N$ | - |
| gfNN (Gutmann and Hyvärinen, 2012) | | ✓ | × | A,L | SGD | H-O | $\left(\widetilde{\boldsymbol{D}}^{-\frac{1}{2}}\widetilde{\boldsymbol{A}}\widetilde{\boldsymbol{D}}^{-\frac{1}{2}}\right)^k$ | $h(\lambda) = (1-\lambda)^k$ |
| SGC (Ribeiro TulioWu and CarlosSingh, 2020) | | ✓ | × | A,L | SGD | H-O | $\left(\widetilde{\boldsymbol{D}}^{-\frac{1}{2}}\widetilde{\boldsymbol{A}}\widetilde{\boldsymbol{D}}^{-\frac{1}{2}}\right)^k$ | $h(\lambda) = (1-\lambda)^k$ |
| ACR-GNN (Peng et al., 2020) | | ✓ | × | A,L | SGD | H-O | - | - |
| RGCN (Yan et al., 2007) | | ✓ | × | A,L | SGD | H-O | $\widetilde{\boldsymbol{D}}^{-\frac{1}{2}}\widetilde{\boldsymbol{A}}\widetilde{\boldsymbol{D}}^{-\frac{1}{2}}$ | - |
| BVAT (Cormen et al., 2001) | | ✓ | × | A,L | Adversarial, SGD | H-O | - | - |
| DropEdge (Pan et al., 2018) | | ✓ | × | A,L | SGD | H-O | $\widehat{\boldsymbol{A}}_{\text{drop}} = \mathcal{N}(\boldsymbol{A} - \boldsymbol{A}')$[105] | - |

(*continued on next page*)

**Table 1** (*continued*)

| Model | Type | Neural | Heter. | A/L | O-S | Proximity | Matrix | Filter |
|---|---|---|---|---|---|---|---|---|
| R-GCN (Bryan et al., 2016) | | ✓ | ✓ | A | SGD | H-O | - | - |
| HetGNN (Wang et al., 2017a) | | ✓ | ✓ | A | SGNS | H-O | - | - |
| GraLSP (Jin et al., 2020) | | ✓ | ✓ | A | SGNS | H-O,ST | - | - |

**Table 2**
Matrices that are implicitly factorized by DeepWalk, LINE and node2vec, same with (Qiu et al., 2018a). "DW" refers to DeepWalk, "n2v" refers to node2vec.

| Model | Matrix |
|---|---|
| DW | $\log\left(\mathrm{vol}(G)\left(\frac{1}{T}\sum_{r=1}^{T}(\boldsymbol{D}^{-1}\boldsymbol{A})^r\right)\boldsymbol{D}^{-1}\right) - \log b$ |
| LINE | $\log(\mathrm{vol}(G)\boldsymbol{D}^{-1}\boldsymbol{A}\boldsymbol{D}^{-1}) - \log b$ |
| n2v | $\log\left(\frac{\frac{1}{2T}\sum_{r=1}^{T}\left(\sum_u \boldsymbol{X}_{w,u}\underline{\boldsymbol{P}}^r_{c,w,u} + \sum_u \boldsymbol{X}_{c,u}\underline{\boldsymbol{P}}^r_{w,c,u}\right)}{(\sum_u \boldsymbol{X}_{w,u})(\sum_u \boldsymbol{X}_{c,u})}\right) - \log b$ |

methods like matrix factorization and spectral embedding models, can be seen in the embedding model ProNE (Zhang et al., 2019b), where vertex embeddings are firstly obtained by factorizing a sparse matrix and then propagated by band-pass filter matrix in the spectral domain. Moreover, such close connections can also be seen int the university of spectral propagation technique proposed in ProNE, which is proved to be a universal embedding enhancement method, improving the quality of vertex embeddings obtained by other shallow embedding models effectively (Zhang et al., 2019b).

Such associations enable some basic ideas of those shallow embedding models can be regarded as the basis of GNN models.

**Heterogeneous Embedding Models.** Based on shallow embedding models, many embedding models for heterogeneous networks can be developed by some techniques, like metapath2vec (Dong et al., 2017), which applies certainty constrictions on the random sampling process and PTE (Tang et al., 2015a) which splits the heterogeneous graph into several homogeneous graphs.

Moreover, various graph content in heterogeneous models, like vertex and edge features and labels evokes the thoughts on how to effectively utilize graph content in the embedding process and also how to become inductive when being applied on dynamic graphs, which is a common feature of real-world graphs. For example, the proposed embedding model GATNE (Cen et al., 2019) applies attention mechanism on vertex features during the embedding process, and try to learn the transformation function applied on vertex contents to make the model become inductive (GATNE-I).

Such design ideas can be seen as basic models for Graph Neural Networks.

**Graph Neural Networks.** Different from above mentioned shallow embedding models, Graph Neural Networks (GNNs) are some kind of deep, inductive embedding models, which can utilize graph contents better and can also be trained with supervised information. The basic idea of GNNs is iteratively aggregating neighbourhood information from vertex neighbours to get a successive view over the whole graph structure.

Based on vanilla GNN models, there is a huge amount of works focusing on developing enhancement techniques (Huang et al., 2018; Yu et al., 2019b; Feng et al., 2020; Hong et al., 2020) to improve the efficiency and effectiveness of GNN models.

Despite the advantages of GNN models, there are also many problems lying in GNN architecture, with also methods proposed to solve such problems, most of which focus on graph regularization (Deng et al., 2019; Verma et al., 2019), basic theories (BarcelEgor et al., 2020), self-supervised learning (Qiu et al., 2020; Hu et al., 2020a), architecture search (Zhou et al., 2019) and so on.

## 4. Shallow embedding models

### 4.1. Neural based

There is a kind of model that is characterized by looking-up embedding tables containing node embeddings as row or column vectors, which are treated as parameters and can be updated during the training process. There are many approaches for updating vertex embeddings. Some extract vertex-context pairs by performing random walks on the graph (e.g., DeepWalk (Bryan et al., 2014), node2vec (Grover and Leskovec, 2016)). They tend to maximize the log-likelihood of observing context vertices for the given target node. These methods are treated as generative models in (Wang et al., 2017a). In generative models, it is assumed that there existing a true connectivity distribution $p_{true}(\cdot|v)$ for each node v and the graph is generated by the connectivity distribution. The co-occurrence frequencies for the vertex-context pairs are then treated as the observed empirical distributions for the underlying connectivity distribution. Some try to model edges directly through the similarities between vertex embeddings of each connected pair (e.g., first-order proximity in LINE (Tang et al., 2015b)) or training a discriminative model (or a classifier) to predict their existence.

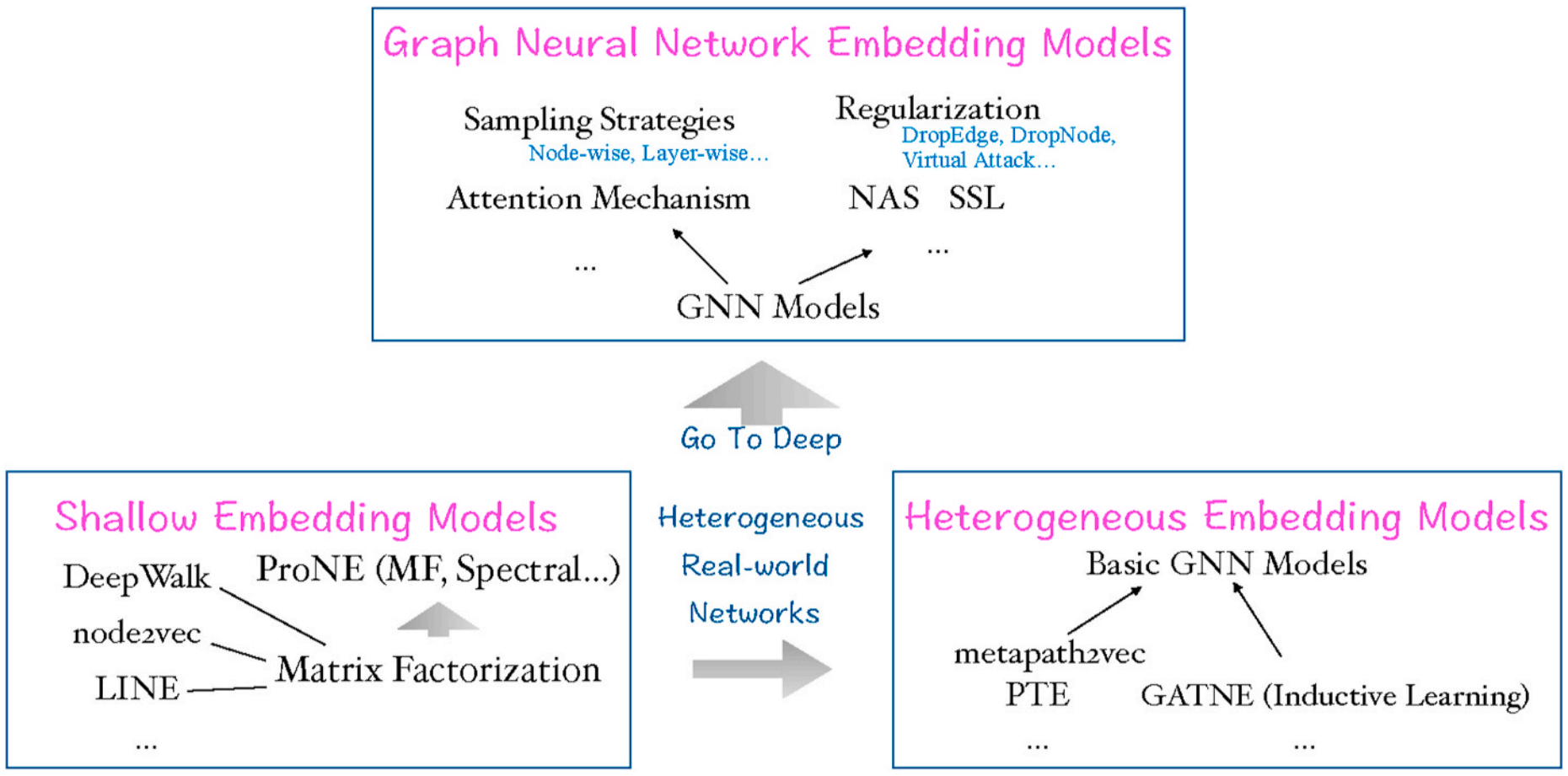

**Fig. 3.** An overview of existing graph embedding models and their correlation.

In this section, we will review a large class of methods based on random walk, make comparisons between different random walk strategies and examine some models adopting other methods.

#### 4.1.1. Random walk family

Random walk and its variants are kind of effective methods transferring the sub-linear structure of graph to the linear structure (i.e., node sequences), since the generated walks can well preserve structural information of the original graph (Micali and Allen, 2016; Lovászet al., 1993).

Random walk strategy is firstly used to generate node sequences in DeepWalk (Bryan et al., 2014), and we refer to the proposed random walk technique as Vanilla Random Walk. It can be seen as a Markov process on the graph and has been well studied (Lovászet al., 1993). Readers can refer to (Lovászet al., 1993) for more details. After node sequences are generated, the Skip-Gram model (Mikolov et al., 2013a) is applied to extract positive vertex-context pairs from them. Based on the distributional hypothesis (Harris, 1954), the Skip-Gram model is first proposed in (Mikolov et al., 2013a) to capture semantic similarity between words in natural language. It is then generalized to networks based on the hypothesis that vertices that share similar structural contexts tend to be close in the embedding space.

**Development of Random Walks.** Based on the vanilla random walk, which is proposed in DeepWalk (Bryan et al., 2014) and has achieved the state-of-the-art performance at that time when applied to downstream tasks (e.g., multi-label node classification), the biased random walk is proposed in (Grover and Leskovec, 2016) by introducing a return parameter p, and a in-out parameter q in the calculation for the transition probability at each step (Fig. 4 Right Channel). Thus, it is also the second-order random walk, whose transition probability also depends on the previous node. Euler walk is proposed in Diff2Vec (Rozemberczki and Sarkar, 2020), which perform a euler tour in the diffusion subgraph centered at each node. Walklets is proposed to separated mixed node proximities information from each order in (Bryan et al., 2016). Thus, it can get embeddings with successively coarser node proximity information preserved as the order k increases. Besides, attribute random walk is proposed in Rol2Vec (AhmedRyan et al., 2018) to design a kind of random walk that can incorporate vertex attributes and structural information.

**Comparison and Discussion.** Compared with the vanilla random walk, the introduced parameters p and q can help the biased random walk interpolate smoothly between DFS and BFS (Cormen et al., 2001). Thus, the biased random walk can explore various node proximities that may exist in the real-world network (e.g., second-order similarity, structural equivalence). It can also fit in a new network more easily by changing parameters to change the preference of proximities being explored since different proximities may dominate in different networks (Grover and Leskovec, 2016). But these two parameters will need tuning to fit in a new graph if there is no labeled data that can be used to learn them.

Both biased random walk and vanilla random walk need calculating transition probabilities for each adjacent node of the current node at each step, which is time-consuming (Rozemberczki and Sarkar, 2020). Compared with them, Euler tour is easy to find in the subgraph (West, 2001). It can also get a more comprehensive view over the neighbourhood since the Euler tour will include all the adjacencies in the subgraph. Thus, fewer diffusion subgraphs and fewer Euler walks need generating centered at each node, compared with vanilla random walks, which tend to revisit a vertex many times, thus producing redundant information (AlonChen et al., 2007; Rozemberczki and Sarkar, 2020). However, the BFS strategy which is used to generate diffusion subgraphs is rather rigid, and cannot explore the various node proximities flexibly. Besides, the effectiveness of Diff2Vec is not well proved, since its performance in popular downstream tasks that are widely used in previous works (e.g., node classification and link prediction) (Grover and Leskovec, 2016; Bryan et al., 2014; Tang et al., 2015b; Donnat et al., 2017; Qiu et al., 2018a; Zhang et al., 2019b) have not been studied (Rozemberczki and Sarkar, 2020).

#### 4.1.2. Others methods

Random walk based methods can be seen as kinds of generative models (Wang et al., 2017a). There are also other methods coming out of the random walk and Skip-Gram range, which can be seen as discriminative models, or as implicitly both generative and discriminative (e.g., LINE (Tang et al., 2015b)), or as adversarial generative training method (Goodfellow et al., 2014) (e.g., GraphGAN (Wang et al., 2017a)).

In LINE, both the existence of edges and the connectivity distribution for each node are modeled, which can be seen as its discriminative and generative parts respectively. Existence of edges is modeled by maximizing following probability for each two connected node pair $(v_i, v_j)$:

$$p_1(v_i, v_j) = \frac{1}{1 + \exp\left(-\vec{u}_i^T \cdot \vec{u}_j\right)}, \tag{1}$$

where $\vec{u}_i$ is the vertex embedding for node $v_i$. The connectivity distribution for each node $p_2(\cdot|v_i)$ (can be calculated by Eq. (3)) is forced to be similar with the empirical distribution $\hat{p}_2(\cdot|v_i)$ by minimizing the following objective to model the second-order proximity:

$$O_2 = \sum_{i \in V} \lambda_i d\left(\hat{p}_2(\cdot|v_i), p_2(\cdot|v_i)\right), \tag{2}$$

where $d(\cdot,\cdot)$ is the distance between two distributions, $\lambda_i$ is the weight for each node, which represents its prestige in the network and can be measured by vertex degree or other algorithms (e.g. PageRank (PAGE, 1998)).

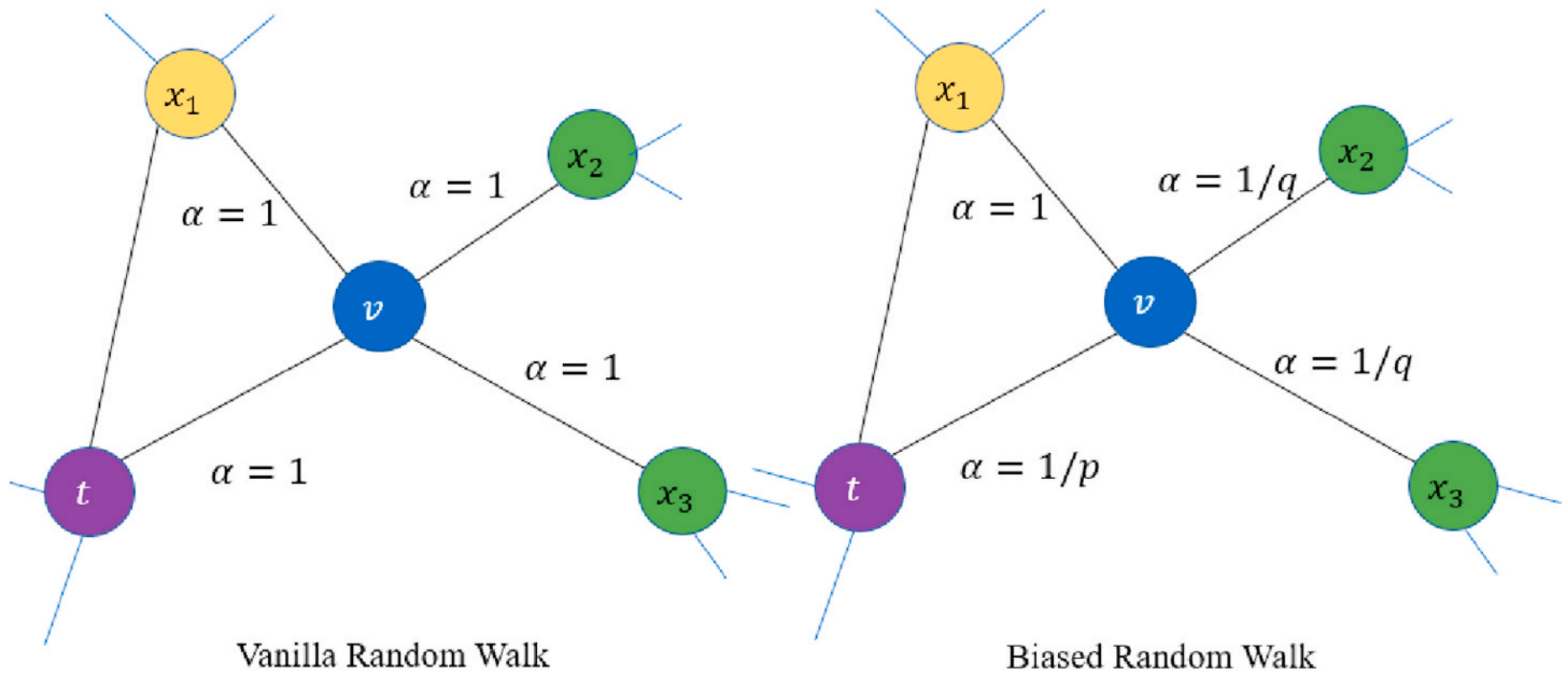

**Fig. 4.** An illustration for the transition probabilities in vanilla and biased random walk. Right Panel: Assuming the previous node is t and the current node is v, then $\alpha_{pa}(v,x)$ for node $x_1, x_2, x_3$ depend on their distances from the previous node t. The transition probability from current node v to node x is calculated by $\pi_{vx} = \alpha_{pq}(t,x) \cdot w_{vx}$, where $w_{vx}$ is the weight of edge $(v,x)$. Left Panel: The vanilla random walk can be regarded as a special case of the biased random walk, where $p = q = 1$. Adapted from (Grover and Leskovec, 2016).

$$p_2(v_j|v_i) = \frac{\exp\left(\overrightarrow{u}_j'^T \cdot \overrightarrow{u}_i\right)}{\sum_{k=1}^{|V|}\exp\left(\overrightarrow{u}_k'^T \cdot \overrightarrow{u}_i\right)} \tag{3}$$

### 4.2. Matrix factorization based

Matrix factorization is an effective method to get high-quality vertex embedding vectors.

Matrices to be factorized can be defined to preserve various node proximities, like the first-order, second-order and intra-community proximities preserved in M-NMF (Wang et al., 2017b), the asymmetric high-order node proximity preserved in HOPE (Ou et al., 2016). Or they can be defined as the matrix implicitly factorized by shallow neural embedding models discussed before, since some of these methods are proved to be inherently related to matrix factorization. It will be discussed in the next subsection.

Moreover, there are many techniques to factorize the matrix. In addition to making the factor matrices obtained by factorizing preserve the properties of the original matrix, time efficiency is also of great importance for matrix factorization methods. Detailed discussion can be seen in Section 8.2.

### 4.3. Connection between neural based and matrix factorization based models

Recent years have seen many works focusing on the exploring the equivalence between some of shallow neural embedding models and matrix factorization models by proving that some neural based models are factorizing matrices implicitly. Such connections can also help with the analysis of robustness of random walk based embedding models (Bojchevski and Günnemann, 2019). Moreover, it is empirically proved that embedding vectors obtained by factorizing the corresponding matrix can preform better in downstream tasks than those optimized by stochastic gradient descent in DeepWalk (Qiu et al., 2018a).

#### 4.3.1. Matrices in natural language models

This concern for equivalence does not originate from graph representation learning models. It is proposed in (Levy and Goldberg, 2014) that the word2vec model (Mikolov et al., 2013b) or the SGNS procedure in it is implicitly factorizing the following word-context matrix:

$$M_{ij}^{SGNS} = \log\left(\frac{\#(w,c)\cdot|D|}{\#(w)\cdot\#(c)}\right) - \log(k), \tag{4}$$

where $\#(w,c)$ is the number of co-occurrence of the word pair $(w,c)$ in the corpus, which is selected by sliding a certain length of window over word sequences, $\#(w)$ is the number of occurrences of the word w. It is worth noting that the term $\log\left(\frac{\#(w,c)\cdot|D|}{\#(w)\cdot\#(c)}\right)$ is actually the well known pointwise mutual information (PMI) of the word pair $(w,c)$ and has been widely used in word embedding models (Church and Hanks, 1990; Ido et al., 1994; Turney, 2001; Turney and Pantel, 2010).

Moreover, PMI is also the basis for deriving the matrices factorized by random walk based models in (Qiu et al., 2018a), which are finally presented in matrix form.

#### 4.3.2. From natural language to graph

For graph representation learning models, some typical algorithms (e.g. DeepWalk (Bryan et al., 2014), node2vec (Grover and Leskovec, 2016), LINE (Tang et al., 2015b)) can also be shown to factorize their corresponding matrices implicitly (Table 2). Based on SGNS's implicit matrix $M^{SGNS}$ (Eq. (4)), the proof focuses on building the bridge between PMI of word-context pair $(w,c)$ and the transition probability matrix of the network.

**Factorizing Log-Empirical-Distribution Matrices.** Theoretical results for the connections between shallow neural embedding algorithms and matrix factorization open a new direction for the optimization process of some neural based methods. Since each entry for this kind of matrices can be seen as the empirical connectivity preference (Wang et al., 2017a) between the corresponding vertex-context pair $(w,c)$, we refer to these matrices as Log-Empirical-Distribution Matrices. In (Qiu et al., 2018a), Qiu et al. try to factorize the matrix of DeepWalk (Bryan et al., 2014) directly. Embedding vectors generated this way can outperform the embedding vectors obtained by the SGNS process employed in the original DeepWalk algorithm in the downstream tasks. In the matrix factorization part of ProNE (Zhang et al., 2019b), a matrix (Eq. (5), where λ is the negative sampling ratio and $P_{D,j}$ are negative samples associated with node $v_j$.) with only the first-order node proximities preserved (thus a sparse one) is generated through a similar way in (Levy and Goldberg, 2014) and is factorized to get raw embedding vectors.

$$M_{i,j} = \begin{cases} \ln p_{i,j} - \ln(\lambda P_{D,j}) & ,(v_i,v_j) \in \mathscr{D} \\ 0 & ,(v_i,v_j) \notin \mathscr{D} \end{cases} \tag{5}$$

In GraRep (Cao et al., 2015), SGNS matrices preserving each k-order proximity:

$$Y_{i,j}^k = \log\left(\frac{A_{i,j}^k}{\sum_t A_{t,j}^k}\right) - \log(\beta), \tag{6}$$

where A is the adjacency matrix, are generated and then factorized to get the embedding vectors preserving each k-order proximities. These embedding vectors, preserving different orders of node proximity, are then concatenated together to get the final embedding vectors for each node.

#### 4.3.3. Differences between neural based embedding and matrix factorization based models

Although SGNS can be shown to implicitly factorize a matrix, there are also many differences between them.

SGNS needs to sample node pairs explicitly, which is time-consuming if we want to preserve high-order node proximities. At the same time, the matrix being generated is also a dense one, if high-order proximites are preserved. But a dense matrix can sometimes be approximated or replaced by a sparse one and then adopt other refinement methods (Zhang et al., 2019b; Qiu et al., 2019). Thus, matrix factorization methods are more likely to be scaled to large-scale networks since complexity for factorizing a sparse matrix can be controlled to $O(|E|)$ with the development of numerical computation (Xu et al., 2018; Zhang et al., 2019b). Besides, factorizing matrices does not require tuning learning rates or other hyper-parameters.

However, factorizing matrices always suffer from unobserved data, which can be weighted naturally in sampling based methods (Levy and Goldberg, 2014). In contrast, exactly weighting for matrix factorization is a hard computational problem.

### 4.4. Enhancing via graph spectral filters

Apart from the shallow neural embedding models and models which adopt matrix factorization to generate vertex embeddings, recent literature has seen the wide application of graph spectral filters in generating high-quality and structure-aware vertex embeddings.

For example, in ProNE (Zhang et al., 2019b), the band-pass filter $g(\lambda) = e^{-\frac{1}{2}[(\lambda-\mu)^2-1]\theta}$ (Shuman et al., 2016; Hammond et al., 2011) is designed to propagate raw vertex embedding vectors generated by factorizing a sparse matrix (Eq. (5)) in the first stage. The propagation operation is also empirically proved to be an effective and universal method that can improve the quality of vertex embedding vectors obtained by many other embedding algorithms (e.g. DeepWalk (Bryan et al., 2014), node2vec (Grover and Leskovec, 2016), GraRep (Cao et al., 2015), HOPE (Ou et al., 2016), LINE (Tang et al., 2015b)).

In the embedding refinement stage of GraphZoom (Deng et al., 2020), it is found that the solution of the refinement problem:

$$\min_{\boldsymbol{E}} \left|\left|\boldsymbol{E}_i - \widehat{\boldsymbol{E}}_i\right|\right|_2^2 + \mathrm{tr}\left(\boldsymbol{E}_i^T \boldsymbol{L}_i \boldsymbol{E}_i\right), \tag{7}$$

where $\widehat{\boldsymbol{E}}_i$ is the embedding matrix to be refined, $\boldsymbol{E}_i$ is the desired matrix after refinement, $\boldsymbol{L}_i$ is the corresponding graph Laplacian, is:

$$\boldsymbol{E}_i = (\boldsymbol{I} + \boldsymbol{L}_i)^{-1}\widehat{\boldsymbol{E}}_i. \tag{8}$$

It is equal to passing the original vertex embeddings through the low-pass filter $h(\lambda) = (1+\lambda)^{-1}$ in the spectral domain. The filter $h(\lambda) = (1+\lambda)^{-1}$ is further approximated by its first-order approximation $\widetilde{h}(\lambda) = 1 - \lambda$ and then generalized to the k-order multiplication form: $\widetilde{h}_k(\lambda) = (1-\lambda)^k$. Its matrix form $\left(\widetilde{\boldsymbol{D}}^{-\frac{1}{2}}\widetilde{\boldsymbol{A}}\widetilde{\boldsymbol{D}}^{-\frac{1}{2}}\right)^k$, where $\widetilde{\boldsymbol{A}}$ is the augmented adjacency matrix, is used to filter the embedding matrix $\widehat{\boldsymbol{E}}_i$ to get the refined embedding matrix $\boldsymbol{E}_i$.

In GraphWave (Donnat et al., 2017), the low-pass filter $g_s(\lambda) = e^{-\lambda s}$ is used to generate the spectral graph wavelet $\Psi_a$ for each node a in the graph:

$$\Psi_a = \boldsymbol{U}\mathrm{diag}(g_s(\lambda_1), \ldots, g_s(\lambda_N))\boldsymbol{U}^T\delta_a, \tag{9}$$

where $\boldsymbol{U}$, $\lambda_1, \ldots, \lambda_N$ are the eigenvector matrix and eigenvalues of the combinational graph Laplacian $\boldsymbol{L}$ respectively, $\delta_a = 1(a)$ is the one hot vector for node a. And m-th wavelet coefficient of this column vector $\Psi_a$ is denoted by $\Psi_{ma}$. By characterizing the distribution via empirical characteristic functions:

$$\Phi_a(t) = \frac{1}{N}\sum_{m=1}^{N} e^{it\Psi_{ma}}, \tag{10}$$

And concatenating $\Phi_a(t)$ at d evenly spaced points $t_1, \ldots, t_d$ as follows (Eq. (11)), a $2d$- dimension embedding vector for node a can be generated:

$$\chi_a = [\mathrm{Re}(\Phi_a(t_i)), \mathrm{Im}(\Phi_a(t_i))]_{t_1, \ldots, t_d}. \tag{11}$$

It can be proved that the k-hop structural equivalent and similar nodes a and b will have ε-structural similar wavelets $\Psi_a$ and $\Psi_b$, where ε is the K-th order polynomial approximation error of the low-pass kernel $g_s(\lambda)$. Thus, the embedding vectors generated by wavelets can preserve the structural similarity.

**Universal Graph Spectral Filters.** Graph spectral filters have close connections with spatial properties. In fact, many models have the corresponding spectral filters as their kernels, which are further discussed in Section 7.1. Readers can refer to (ShumanSunil et al., 2013; Von Luxburg, 2007; Hammond et al., 2011; Tremblay and Borgnat, 2014; David Shuman et al., 2013) for more details.

## 5. Heterogeneous embedding models

Although the embedding models discussed above are designed for homogeneous networks, they are actually the basis of many heterogeneous networks.

Heterogeneous networks are widespread in real-world, which have more than one type of vertices or edges. Thus, algorithms for heterogeneous network embedding are supposed to not only incorporate vertex attributes or labels with structural information, but also leverage vertex types, edge types, and also the semantic information that lies behind the connections between two vertices (Dong et al., 2020). This is exactly where the challenge of heterogeneous network representation learning lies in.

Since there are already surveys for heterogeneous networks representation learning algorithms (Dong et al., 2020; Yang et al., 2020a), we will focus on the correlations between heterogeneous and homogeneous network embedding techniques in this section.

### 5.1. Heterogeneous LINE

In PTE (Tang et al., 2015a), the heterogeneous network that has words, documents, labels as its vertices and the connections within them as the edges, is projected to three homogeneous networks first (word--word network, word-document network and word-label network). Then, for each bipartite network $G = (\mathscr{V}_A \cup \mathscr{V}_B, \mathscr{E})$, where $\mathscr{V}_A$ and $\mathscr{V}_B$ are two disjoint vertex sets, $\mathscr{E}$ is the edge set, the conditional probability of vertex $v_j$ in set $\mathscr{V}_A$ generated by vertex $v_i$ in set $\mathscr{V}_B$ is defined as:

$$p\left(v_j|v_i\right) = \frac{\exp\left(\overrightarrow{u}_j^T \cdot \overrightarrow{u}_i\right)}{\sum_{k\in A}\exp\left(\overrightarrow{u}_k^T \cdot \overrightarrow{u}_i\right)}, \tag{12}$$

Similar with $p_2(v_j|v_i)$ (Eq. (3)) in LINE (Tang et al., 2015b). Then the conditional distribution $p(\cdot|v_j)$ is forced to be close to its empirical distribution $\widehat{p}(\cdot|v_j)$ by jointly minimizing the corresponding loss function similar with the one in LINE (Eq. (2)).

Moreover, it is also proved in (Qiu et al., 2018a) that the implicit matrix factorized by PTE is in the following form:

$$\log\left(\begin{bmatrix} \alpha \mathrm{vol}(G_{\mathrm{ww}})\left(\boldsymbol{D}_{\mathrm{row}}^{\mathrm{ww}} - 1\right)\boldsymbol{A}_{\mathrm{ww}}\left(\boldsymbol{D}_{\mathrm{col}}^{\mathrm{ww}} - 1\right) \\ \beta \mathrm{vol}(G_{\mathrm{dw}})\left(\boldsymbol{D}_{\mathrm{row}}^{\mathrm{dw}} - 1\right)\boldsymbol{A}_{\mathrm{dw}}\left(\boldsymbol{D}_{\mathrm{col}}^{\mathrm{ww}} - 1\right) \\ \gamma \mathrm{vol}(G_{\mathrm{lw}})\left(\boldsymbol{D}_{\mathrm{row}}^{\mathrm{lw}} - 1\right)\boldsymbol{A}_{\mathrm{lw}}\left(\boldsymbol{D}_{\mathrm{col}}^{\mathrm{ww}} - 1\right) \end{bmatrix}\right) - \log \boldsymbol{b}, \tag{13}$$

where $G_{ww}, G_{dw}, G_{lw}$ are word-word, document-word, label-word graphs respectively, with $A_{ww}, A_{dw}, A_{lw}$ as their adjacency matrices and $D^{ww}$, $D^{dw}, D^{lw}$ as their degree matrices respectively, $\mathrm{vol}(G) = \Sigma_i\Sigma_j A_{ij} = \Sigma_i d_i$ is the volume of the weighted graph G.

### 5.2. Heterogeneous random walk

The proposed meta-path based random walk in (Dong et al., 2017) provides a natural way to transform the heterogeneous networks into vertex sequences with both structural information and semantic information underlying different types of vertices and edges preserved. The key idea is to design specific meta paths which can restrict transitions between only specified types of vertices. To be specific, given a heterogeneous network $G = (\mathscr{V}, \mathscr{E})$ and a meta path scheme $\mathscr{P}: V_1 R_1 \rightarrow V_2 R_2 \rightarrow V_3 \cdots V_t R_t \rightarrow \cdots R_{l-1} \rightarrow V_l$, where $V_i \in \mathscr{O}$ are vertex types in the network, the transition probability is defined as:

$$P_{v^{i+1}, v_t^i, \mathscr{P}} = \begin{cases} \frac{1}{\left|\mathscr{N}_{t+1}\left(v_t^i\right)\right|} & \left(v^{i+1}, v_t^i\right) \in \mathscr{E}, \Phi\left(v^{i+1}\right) = t+1 \\ 0 & \left(v^{i+1}, v_t^i\right) \in \mathscr{E}, \Phi\left(v^{i+1}\right) \neq t+1 \\ 0 & \left(v^{i+1}, v_t^i\right) \notin \mathscr{E} \end{cases} \tag{14}$$

where $\Phi(v_t^i) = V_t$, $\mathscr{N}_{t+1}(v_t^i)$ is the $V_{t+1}$ type of neighbourhood of vertex $v_t^i$. Then the SGNS framework is applied to the generated random walks to optimize vertex embeddings. Moreover, the type-dependent negative sampling strategy is also proposed to better capture the structural and semantic information in heterogeneous networks.

**Combined with Neighbourhood Aggregation.** Different from the paradigm where node embedding and edge embedding vectors are defined directly as parameters to be optimized, attention mechanism is used in GATNE (Cen et al., 2019) to calculate embedding vectors for each node based on neighbourhood aggregation operation. Two embedding methods are introduced: GATNE-T (transductive) and GATNE-I (inductive).

In GATNE, the overall embedding of node $v_i$ on edge r is split into base embedding which is shared between different edge types and edge embedding. The k-th level edge embedding $u_{i,r}^{(k)} \in \mathbb{R}^s$, $(1 \le k \le K)$ of node $v_i$ on edge type r is aggregated from neighbours' edge embeddings:

$$u_{i,r}^{(k)} = aggregator\left(u_{j,r}^{(k-1)}\right), \forall v_j \in \mathcal{N}_{i,r}, \tag{15}$$

where $\mathcal{N}_{i,r}$ is the neighbours of node $v_i$ on edge type r. After the K-th level edge embeddings are calculated, the overall embedding $v_{i,r}$ of node $v_i$ on edge type r is computed by applying self-attention mechanism on the concatenated embedding vector of node $v_i$:

$$U_i = \left(u_{i,1}, u_{i,2}, \ldots, u_{i,m}\right). \tag{16}$$

The base embedding $b_i$ for node $v_i$ is then added as the embedding bias on the self-attention result.

The difference between transductive model (GATNE-T) and the inductive model (GATNE-I) lies in how the 0-th level edge embedding vector of each node $v_i$ on each edge type r and the base embedding vector of each node $v_i$ is calculated. In GATNE-T, they are optimized directly as parameters, while in GATNE-I, they are computed by applying transformation functions $h_z$ and $g_{z,r}$ on the raw feature $x_i$ of each node $v_i$: $b_i = h_z(x_i), u_{i,r}^{(0)} = g_{z,r}(x_i)$. Transformation functions $h_z$ and $g_{z,r}$ are optimized during the training process.

The neighbourhood aggregation mechanism increases the model's inductive bias and also make it easier combining with node features, which is similar with the core idea of inductive embedding models based on graph neural networks.

**Meta Paths Augmentation.** Meta-paths can also be treated as the relations or "edges" between the corresponding connected vertices to augment the network. In HIN2vec (Fu et al., 2017), meta-paths are treated as the relations between vertices connected by them with learnable embeddings. Then probability of the two vertices x and y connected by meta-path r is modeled by:

$$P(r|x,y) = \text{sigmoid}\left(\sum \boldsymbol{W}_X{}' \vec{x} \odot \boldsymbol{W}_Y{}' \vec{y} \odot f_{01}(\boldsymbol{W}_R{}' \vec{r})\right), \tag{17}$$

where $\boldsymbol{W}_X$, $\boldsymbol{W}_Y$ are vertex embedding matrices, $\boldsymbol{W}_R$ is the relation embedding matrix, $\boldsymbol{W}_X{}'$ is the transpose of matrix $\boldsymbol{W}_X$, $\vec{x}, \vec{y}, \vec{r}$ are one-hot vectors for two connected vertices x, y and the relation between them respectively, $f_{01}(\cdot)$ is the regularization function. Parameters are optimized by maximizing the following objective:

$$\begin{aligned} &\log O_{x,y,r}(x,y,r) = L(x,y,r)\log P(r|x,y) + \\ &(1 - L(x,y,r))\log(1 - P(r|x,y)), \end{aligned} \tag{18}$$

where $L(x,y,r) = 1$ if vertices $x, y$ is connected by relation r, otherwise $L(x,y,r) = 0$.

In TapEM (Chen and Sun, 2017), the proximity between two vertices $i, j$ on two sides of the given meta-path of type r is explicitly preserved by modeling the conditional probability $P(j|i;r)$, where $i, j$ are vertices on two sides of the meta-path, r is the type of the meta-path.

In HeteSpaceyWalk (Yu et al., 2019a), the heterogeneous personalized spacey random walk algorithm is proposed, which is a space-friendly and efficient approximation for meta-path based random walks and can converge to the same limiting stationary distribution.

**Summary.** Compared with PTE, random walk for heterogeneous networks can capture the structural dependencies between different types of vertices better and also preserve higher-order proximities. But they both need manual design with expert knowledge in advance (how to separate networks in PTE and how to design meta paths). Moreover, just using the information of types of the meta path between two connected vertices may lose some information (e.g., vertex or edge types, vertex attributes) passing through the meta path (Hong et al., 2020).

### 5.3. Other models

Apart from the random walks and Skip-Gram framework, there are also many other homogeneous embedding models that can be used in heterogeneous network embedding algorithms (e.g., label propagation, matrix factorization, generate adversarial approach applied in GraphGAN (Wang et al., 2017a)). The assumption of label propagation in homogeneous networks that "two connected vertices tend to have the same labels" is generalized to the heterogeneous networks in LSHM (Jacob et al., 2014) by assuming that two vertices of the same type connected by a path tend to have similar latent representations. GAN (Goodfellow et al., 2014) is used in HeGAN (HuYuan and Shi, 2019) with relation-aware discriminator and generator to perform better negative sampling.

Moreover, supervised information can also be added for downstream tasks. For example, the idea of PTE, meta paths and matrix factorization are combined in HERec (Shi et al., 2019) with supervised information from recommendation task. To be specific, vertex and item embedding vectors are firstly generated by performing meta path-based random walks on the graph and then the user-item rating matrix is introduced to help learn fusion functions on those embeddings.

## 6. Graph neural network based models

Graph neural networks (GNNs) are kind of powerful feature extractor for graph structured data and have been widely used in graph embedding problems. There are some inherent problems in shallow embedding models, which will be discussed in later sections, and the presence of GNNs can alleviate these problems to some extent.

Past few years have seen the rapid development of Graph Neural Networks in graph mining tasks. GNNs' architecture can enable them to effectively model structural and relational data. Compared with shallow embedding models that have been discussed before, GNNs have a deep architecture and can model vertex attributes as well as network structure naturally. These are typically neglected, or cannot be modeled efficiently in shallow embedding models.

There are two main streams in designing GNNs. The first is in the graph spectral fashion, in which the convolutional operation can be seen as passing vertex features through a low-pass filter in the spectral domain. We refer to these GNNs as graph spectral GNNs. The classical graph convolution network (GCN) (Kipf and Welling, 2016, 2017), ChebyNet (Defferrard et al., 2016), FastGCN (Chen et al., 2018), ASGCN (Huang et al., 2018), GWNN (Xu et al., 2019a) and the graph filter network (gfNN) proposed in (Hoang and Maehara, 2019) are examples of graph spectral GNNs. The convolution process is performed in the spectral domain in such GNNs, in which node features are first transferred to spectral domain and then multiply with a spectral filter matrix. Desired spectral convolution process should be economic in computation and also localized in spatial domain (Defferrard et al., 2016; Xu et al., 2019a). Then, the second type is graph spatial GNNs, which operate on vertex features in the spatial domain directly. They update each node's features by linearly combining (or aggregating) its neighbours' features. It is similar to the spatial convolution discussed in (Hammond et al., 2011). GraphSAGE (Hamilton et al., 2017), Graph Isomorphism Network (GIN) (Xu et al., 2019b), and MPNN (Gilmer et al., 2017) are examples for this kind of GNNs.

Compared with spectral GNNs, the spatial convolution employed in the spatial GNNs usually just focus on 1-st neighbours of each node. However, the local property of spatial convolution operation can help spatial GNNs be inductive.

Compared with shallow embedding models, GNNs can better combine the structural information with vertex attributes, but the need for vertex attributes also make GNNs hard to be applied to homogeneous networks without vertex features. Although it has been proposed in (Kipf and Welling, 2017) that we can set the feature matrix $\boldsymbol{X} = \boldsymbol{I}_N$, where $\boldsymbol{I}_N \in \mathbb{R}^{N \times N}$ is the identity matrix and N is the number of vertices, for featureless graphs, it cannot be scaled to large networks.

Apart from GNNs' advantages in content augmentation, they can also be trained in the supervised or semi-supervised fashion easily (in fact, GCN is proposed for semi-supervised classification). Label augmentation can improve the discriminative property of the learned features (Zhang et al., 2018). Moreover, neural network architecture can help with the design of an end-to-end model, fitting in downstream tasks better.

### 6.1. GNN models

While the powerful CNNs can also be effectively applied to data that can be organized in grid structures (e.g. images (Krizhevsky et al., 2012), sentences (Yoon, 2014), and videos (Taylor et al., 2010)), they cannot be generalized to graphs directly. Inspired by the effectiveness of CNNs in extracting features from grid structures, many previous works focus on properly defining the convolution operation for graph data to capture structural information.

To the best of our knowledge, it was in (Bruna et al., 2013) that convolutions for graph data were first introduced based on graph spectral theory (Defferrard et al., 2016) and graph signal processing (ShumanSunil et al., 2013), where both multilevel convolutional neural networks in spectral and spatial domains were built with few parameters to learn, which preserve nice qualities for CNNs. The spectral convolution operation in (Bruna et al., 2013) is actually a low-pass filtering operation, consistent with the ideas for building graph neural networks in the following works (Kipf and Welling, 2017; Hamilton et al., 2017; Veličković et al., 2018; Hoang and Maehara, 2019).

GCN is proposed in (Kipf and Welling, 2017), which uses the first-order approximation of the graph spectral convolution and the augmented graph adjacency matrix to design the feature convolution layer's architecture.

After the proposition of GCN (Kipf and Welling, 2017), many GNN models are designed based on it. They try to make some improvements, such as introducing sampling strategies (Hamilton et al., 2017; Huang et al., 2018; Chen et al., 2018), adding attention mechanism (Veličković et al., 2018; Kiran et al., 2018), or improving the filter kernel (Wu et al., 2019; Hoang and Maehara, 2019). We briefly summarize parts of existing GNN models in Table 3 and discuss some examples in the following parts.

#### 6.1.1. Sampling

Sampling techniques are introduced to reduce the time complexity of GCN or introduce the inductive bias. There are various sampling strategies and we will introduce some of them in the following parts, such as node-wise sampling (Hamilton et al., 2017; Jin et al., 2020), layer-wise sampling (Chen et al., 2018; Huang et al., 2018) and subgraph sampling (Zeng et al., 2020).

**Node-Wise Sampling.** Based on GCN, GraphSAGE (Hamilton et al., 2017) introduces node-wise sampling to randomly sample a fixed size neighbourhood for each node in each layer and also shift to the spatial domain to help it become inductive.

However, as proposed in ASGCN (Huang et al., 2018), its node-wise sampling strategy would probability lead to the number of sampled nodes grows exponentially with the number of layers. If the depth of the network is d, then the number of sampled nodes in the input layer will increase to $O(n^d)$, where n is the number of sampled neighbours of each node, leading to significant computational burden for large d.

**Table 3**
A summary of Graph Neural Networks. Part of the symbols in the formula can refer to Def. 5. For others, $\boldsymbol{H}^{(l)}$ denotes to the feature matrix in layer l, $h_v^k$ denotes feature vector of node v in layer k, $h^{(l)}(v_i)$ or $h_i^{(l)}$ denotes feature vector of node $v_i$ (or i) in layer l, $\boldsymbol{\Theta}$ and $\boldsymbol{W}$ are trainable parameters, $\boldsymbol{M}^{(l)}$ and $\boldsymbol{\Sigma}^{(l)}$ are mean and variance matrices of vertex features in layer l respectively, $\mathcal{N}(v)$ denotes the set of node v's neighbours or the sampled neighbours, $h^k_{\mathcal{N}(v)}$ denotes the feature vector aggregated from node v's sampled neighbours in layer k, $q(\cdot)$ denotes the sampling distribution. For abbreviations used, "E" denotes "Spectral", "A" denotes "Spatial", "A-M" denotes "Attention Mechanism", "T" denotes "Type", "S" denotes "Sampling Strategy", "N" refers to "Node-Wise Sampling", "L" refers to "Layer-Wise Sampling".

| Model | T | S | A-M | Aggregation Function | Appendix |
|---|---|---|---|---|---|
| GCN (Kipf and Welling, 2017) | E | × | × | $\boldsymbol{H}^{(l+1)} = \widetilde{\boldsymbol{D}}^{-\frac{1}{2}}\widetilde{\boldsymbol{A}}\widetilde{\boldsymbol{D}}^{-\frac{1}{2}}\boldsymbol{H}^{(l)}\Theta$ | $\widetilde{\boldsymbol{A}} = \boldsymbol{A} + \boldsymbol{I}_N$ |
| GraphSAGE (Hamilton et al., 2017) | E | N | × | $h^k_{\mathcal{N}(v)} \leftarrow \text{AGGREGATE}_k(\{h_u^{k-1}, \forall u \in \mathcal{N}(v)\}) h_v^k \leftarrow \sigma(W^k \cdot \text{CONCAT}(h_v^{k-1}, h^k_{\mathcal{N}(v)}))$ | $\text{AGGREGATE} \in \{\text{MAX POOL}, \text{MEAN}, \text{LSTM}\}$ |
| FastGCN (Chen et al., 2018) | E | L | × | $\boldsymbol{H}^{(l+1)}(v) = \sigma\left(\frac{1}{t_l}\sum_{j=1}^{t_l}\frac{\widehat{A}(v, u_j^{(l)})\boldsymbol{H}^{(l)}(u_j^{(l)})\boldsymbol{W}^{(l)}}{q(u_j^{(l)})}\right),$ <br> $u_j^{(l)} \sim q, l = 0, 1, \ldots, M (19)$ | $q(u) = \frac{\Vert \widehat{A}(:, u)\Vert^2}{\sum_{u' \in V}\Vert \widehat{A}(:, u')\Vert^2},$ <br> $u \in V$ |
| ASGCN (Huang et al., 2018) | E | L | × | $h^{(l+1)}(v_i) = \sigma_{W^{(l)}}\left(N(v_i)\frac{1}{n}\sum_{j=1}^{n}\frac{p(\widehat{u}_j \vert v_i)}{q(\widehat{u}_j \vert v_1, \ldots, v_n)}h^{(l)}(\widehat{u}_j)\right)$ <br> $\widehat{u}_j \sim q(\widehat{u}_j \vert v_1, \ldots, v_n) (20)$ | Eq. (26); $g(x(u_j)) = \boldsymbol{W}_g x(u_j)$ |
| GAT (Veličković et al., 2018) | A | × | ✓ | $h_i^{(l+1)} = CONCAT_{k=1}^K\left[\sigma\left(\sum_{j \in \mathcal{N}_i}\alpha_{ij}^k \boldsymbol{W}^k h_j^{(l)}\right)\right]$ | Eq. (27) |
| RGCN (Zhu et al., 2019) | E | × | × | $\boldsymbol{M}^{(l+1)} = \rho\left(\widetilde{\boldsymbol{D}}^{-\frac{1}{2}}\widetilde{\boldsymbol{A}}\widetilde{\boldsymbol{D}}^{-\frac{1}{2}}(\boldsymbol{M}^{(l)} \odot \mathcal{A}^{(l)})\boldsymbol{W}_\mu^{(l)}\right)$ <br> $\boldsymbol{\Sigma}^{(l+1)} = \rho(\widetilde{\boldsymbol{D}}^{-1}\widetilde{\boldsymbol{A}}\widetilde{\boldsymbol{D}}^{-1}(\boldsymbol{\Sigma}^{(l)} \odot \mathcal{A}^{(l)} \odot \mathcal{A}^{(l)})\boldsymbol{W}_\sigma^{(l)})$ | $\mathcal{A}^{(l)} = \exp(-\gamma\boldsymbol{\Sigma}^{(l)})$ |
| SGC (Wu et al., 2019) | E | × | × | $Y_{SGC} = \text{softmax}\left(\left(\widetilde{\boldsymbol{D}}^{-\frac{1}{2}}\widetilde{\boldsymbol{A}}\widetilde{\boldsymbol{D}}^{-\frac{1}{2}}\right)^K \boldsymbol{X\Theta}\right)$ | $\widetilde{\boldsymbol{A}} = \boldsymbol{A} + \boldsymbol{I}_N$ |
| GIN (Xu et al., 2019b) | A | × | × | $h_v^{(l+1)} = \text{MLP}^{(l+1)}\left((1 + \varepsilon^{(l+1)}) \cdot h_v^{(l)} + \sum_{u \in \mathcal{N}(v)} h_u^{(l)}\right)$ | – |
| ACR-GNN (BarcelEgor et al., 2020) | A | × | × | Eq. (28) | – |
| R-GCN (Schlichtkrull and KipfPeter Bloemvan den Rianne Berg, 2018) | A | × | × | $h_i^{(l+1)} = \sigma\left(\sum_{r \in \mathcal{R}}\sum_{j \in \mathcal{N}_i^r}\frac{1}{c_{i,r}}\boldsymbol{W}_r^{(l)}h_j^{(l)} + \boldsymbol{W}_0^{(l)}h_i^{(l)}\right)$ | $\boldsymbol{W}_r^{(l)} = \sum_{b=1}^{B} a_{rb}^{(l)} V_b^{(l)}$ or <br> $\boldsymbol{W}_r^{(l)} \oplus_{b=1}^{B} \boldsymbol{Q}_{br}^{(l)}$ |
| GraLSP (Jin et al., 2020) | A | N | ✓ | $a_i^{(k)} = \text{MEAN}_{w \in \mathcal{W}^{(i)}, p \in [1, r_w]}(\lambda_{i,w}^{(k)}(q_{i,w}^{(k)} \odot h_{w_p}^{(k-1)}))$ <br> $h_i^{(k)} = \text{ReLU}(\boldsymbol{U}^{(k)}h_i^{(k-1)} + \boldsymbol{V}^{(k)}a_i^{(k)}), k = 1, 2, \ldots, K$ <br> $h_i = h_i^{(K)} (21)$ | $\lambda_{i,w}$: attention coefficient <br> $q_{i,w}$: amplification coefficient <br> $r_w$: receptive window, Eq. (24) |

Different from the random sampling strategy introduced in GraphSAGE, an adaptive node-wise sampling strategy is proposed in GraLSP (Jin et al., 2020). In GraLSP, the vertex v's neighbourhood is sampled by performing random walks of length l starting at vertex v. The aggregation process is also combined with attention mechanism. Structural information is preserved by introducing Random Anonymous Walk (Micali and Allen, 2016), which is calculated based on the sampled random walk $\boldsymbol{w} = (w_1, w_2, \ldots, w_l)$:

$$\mathrm{aw}(\boldsymbol{w}) = (\mathrm{DIS}(\boldsymbol{w}, w_1), \mathrm{DIS}(\boldsymbol{w}, w_2), \ldots, \mathrm{DIS}(\boldsymbol{w}, w_l)), \tag{22}$$

where $\mathrm{DIS}(\boldsymbol{w}, w_i)$ denotes the number of distinct nodes in $\boldsymbol{w}$ when $w_i$ first appears in $\boldsymbol{w}$:

$$\mathrm{DIS}(\boldsymbol{w}, w_i) = \left|\{w_1, w_2, \ldots, w_p\}\right|, p = \min_j\{w_j = w_i\}. \tag{23}$$

Then the adaptive receptive radius is defined as:

$$r_w = \left\lfloor \frac{2l}{\max(\mathrm{aw}(\boldsymbol{w}))} \right\rfloor, \tag{24}$$

where $\max(\mathrm{aw}(\boldsymbol{w}))$ is equal to the number of distinct nodes visited by walk $\boldsymbol{w}$. Then, the first $r_w$ nodes in the random walk $\boldsymbol{w}$ started at node v are chosen to pass their features to node v.

**Layer-Wise Sampling.** Different from node-wise sampling strategies, nodes in the current layer are sampled based on all the nodes in the previous layer. To be specific, layer-wise sampling strategies aim to find the best and tractable sampling distribution $q(\cdot|v_1, \ldots, v_{t_{l-1}})$ for each layer l based on nodes sampled in layer $l-1$: $\{v_1, \ldots, v_{t_{l-1}}\}$. The sampling distributions aim to minimize the variance introduced by performing sampling. However, the best sampling distributions are always cannot be calculated directly, thus some tricks and relaxations are introduced to obtain sampling distributions that can be used in practice (Chen et al., 2018; Huang et al., 2018). The sampling distribution is

$$q(u) = \|\widehat{A}(:, u)\|^2 \Big/ \sum_{u' \in V} \|\widehat{A}(:, u')\|^2, u \in V, \tag{25}$$

in FastGCN (Chen et al., 2018) and

$$q^*(u_j) = , \frac{\sum_{i=1}^{n} p(u_j|v_i)|g(x(u_j))|}{\sum_{j=1}^{N}\sum_{i=1}^{n} p(u_j|v_i)|g(x(u_j))|} \tag{26}$$

in ASGCN (Huang et al., 2018), where $g(x(u_j))$ is a linear function (i.e., $g(x(u_j)) = W_g x_{u_j}$) applied on vertex features of node $u_j$.

**Subgraph Sampling.** Apart from node-wise and layer-wise sampling strategies, which sample a set of nodes in each layer, a subgraph sampling strategy is proposed in (Zeng et al., 2020), which samples a set of nodes and edges in each training epoch and perform the whole graph convolution operation on the sampled subgraph.

Edge sampling rates are proposed aiming to reduce the variance of node features in each layer introduced by performing subgraph sampling strategy. It is set to $p_{u,v} \propto \frac{1}{\deg(u)} + \frac{1}{\deg(v)}$ in practice. Based on edge sampling rates, different samplers can be designed to sample the subgraph. For random edge sampler, edges are sampled just using the edge sampling distribution discussed above. For random node sampler, a certain number of nodes are sampled under the node sampling rate $P(u) \propto \widetilde{A}_{:,u}^{\ 2}$, where $\widetilde{A}$ is the normalized graph adjacency matrix. For random walk based sampler, the sampling rate for the node pair $(u, v)$ is set to $p_{u,v} \propto B_{u,v} + B_{v,u}$, where $\boldsymbol{B} = \widetilde{\boldsymbol{A}}^L$ and L is the length of the random walk. Besides, there are also many other random walk based samplers proposed in previous literature (Bruno Ribeiro and Donald Towsley, 2010; Leskovec and Faloutsos, 2006; Hu and Lau, 2013), which can also be used to sample the subgraph.

### 6.1.2. Attention mechanism

Introducing an attention mechanism can help improve models' capacities and interpretability (Veličković et al., 2018) by assigning different weights to nodes in a same neighbourhood explicitly. It is interesting that the attention mechanism used in GAT (Veličković et al., 2018) will make the model easy to be attacked due to the aggregation process's dependency on neighbours' features. But the one used in RGCN can help improve model's robustness by assigning features with a larger variance lower weights.

Besides, the comparison between attention mechanism and the sampling and LSTM-aggregation strategy used in GraphSAGE can cast some similar insights with the comparison between RNN based models and attention based models for sequence modeling in NLP domain. Attention mechanism can obtain a more comprehensive view over nodes' neighbourhood than RNN based aggregation strategies.

Performing attention mechanism on neighbours' features can also be seen as a feature rescaling process, which can be used to unify GNN models in a same framework. The choice of specific attention strategy depends on our purpose and practice.

In GAT (Veličković et al., 2018), the calculation for k-th head's attention weight $\alpha_{ij}^k$ between two nodes i and j is Eq. (27), where $\overrightarrow{h}_i \in \mathbb{R}^{d\times 1}$ is the feature vector for vertex i, $\boldsymbol{W}^k \in \mathbb{R}^{d\times d'}$, $\overrightarrow{\alpha} \in \mathbb{R}^{2d'\times 1}$ are corresponding parameters, $d, d'$ are the dimension for feature vector in the previous layer and current layer respectively. Different from GAT, the attention weight between nodes i and j is calculated based on the cosine similarity between their hidden representations (Eq. (28), where $\beta(l)$ are trained attention-guided parameters of layer l.) in (Kiran et al., 2018), where $\cos(\cdot,\cdot)$ represents the cosine similarity.

$$\alpha_{ij}^k = \frac{\exp\left(\mathrm{LeakyReLU}\left(\overrightarrow{a}^T\left[\boldsymbol{W}^k\overrightarrow{h}_i \,\|\, \boldsymbol{W}^k\overrightarrow{h}_j\right]\right)\right)}{\sum_{k\in\mathcal{N}_i}\exp\left(\mathrm{LeakyReLU}\left(\overrightarrow{a}^T\left[\boldsymbol{W}^k\overrightarrow{h}_i \,\|\, \boldsymbol{W}^k\overrightarrow{h}_k\right]\right)\right)} \tag{27}$$

$$\alpha_{ij} = \frac{\exp(\beta(l)\cos(H_i, H_j))}{\sum_{j\in\mathcal{N}_i}\exp(\beta(l)\cos(H_i, H_j))} \tag{28}$$

Besides, attention mechanism is also widely used in heterogeneous network embedding algorithms, by applying which the various semantic information underlying different kind of connections between vertices. More discussions can be seen in Section 6.1.4.

### 6.1.3. Discriminative power

**Weisfeiler-Lehman (WL) Graph Isomorphism Test.** GNN's inner mechanism is similar with the Weisfeiler-Lehman (WL) graph isomorphism test (Fig. 5) (Xu et al., 2019b; Hamilton et al., 2017; Shervashidze et al., 2011; Weisfeiler and Lehman, 1968), which is a powerful test (Shervashidze et al., 2011) known to distinguish a broad class of graphs, despite of some corner cases. Comparisons between GNNs and WL test allow us to understand the capabilities and limitations of GNNs more clearly.

It is proved in (Xu et al., 2019b) that GNNs are at most as powerful as the WL test in distinguishing graph structures and can be as powerful as WL test only if using proper neighbour aggregation functions and graph readout functions ((Xu et al., 2019b) Theorem 3). Those functions are applied on the set of neighbours' features, which can be treated as a multi-set (Xu et al., 2019b). For neighbourhood aggregation functions, it is concluded that other multi-set functions like mean, max aggregators are not as expressive as the sum aggregator (Fig. 6).

One kind of powerful GNNs is proposed by taking "SUM" as its aggregation function over neighbours' feature vectors and MLP as its transformation function, whose feature updating function in the k-th

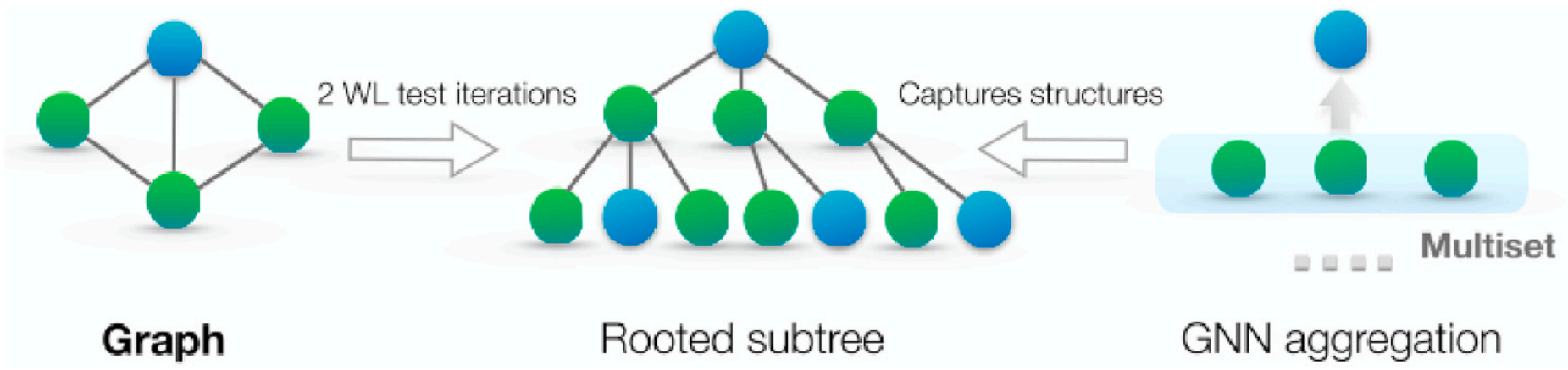

**Fig. 5.** Illustration for WL test and the relationship with GNNs. Middle Panel: rooted subtree of the blue node in the left panel. Right Panel: if a GNN's aggregation function can capture the full multiset of node neighbours, then it can capture the rooted subtree and be as powerful as WL test in distingusihing different graphs. Reprinted from (Xu et al., 2019b).

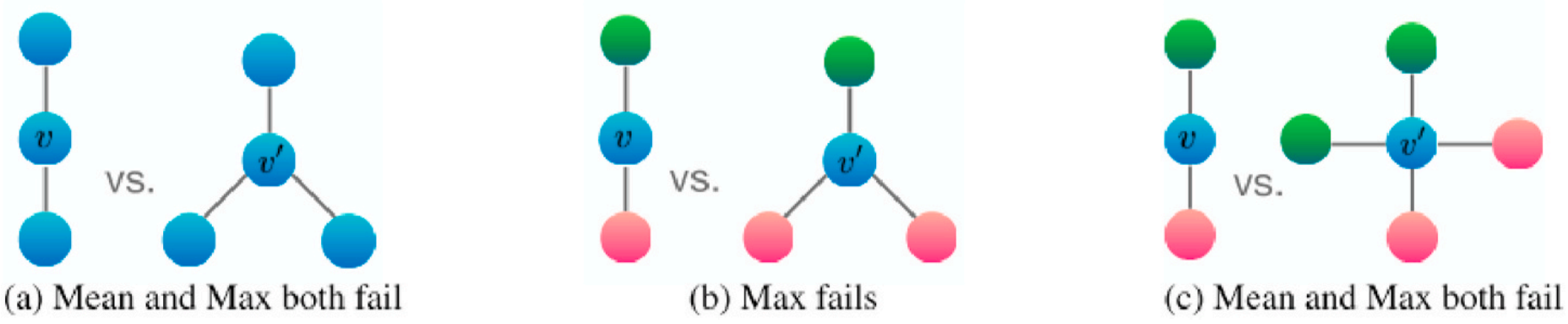

**Fig. 6.** Examples where max aggregator and mean aggregator will fail. For each subimage, node v and node $v'$ will get the same embeddings under corresponding aggregators even though their neighbourhood structures are different from each other. Reprinted from (Xu et al., 2019b).

layer is:

$$h_v^k = MLP^k\left(\left(1+\varepsilon^k\right)\cdot h_v^{k-1} + \sum_{u\in\mathcal{N}(v)} h_u^{k-1}\right). \tag{29}$$

**Logical Classifier.** In (BarcelEgor et al., 2020), Boolean classifiers expressible as formulas in the logic $FOC_2$, which is a well-studied fragment of first-order logic, are studied and used to judge the GNNs' logical expressiveness. It is shown that a popular class of GNNs, called AC-GNNs (Aggregate-Combine GNNs, whose feature updating function can be written as Eq. (30), where COM = COMBINE, AGG = AGGREGATE, $x_v^{(i)}$ is the feature vector of vertex v in layer i) in which the features of each node in the successive layers are only updated in terms of node features of its neighbourhood, can only capture a specific part of $FOC_2$ classifiers.

$$x_v^{(i)} = \mathrm{COM}^{(i)}\left(x_v^{(i-1)}, \mathrm{AGG}^{(i)}\left(\left\{x_u^{(i-1)} \middle| u\in\mathcal{N}_G(v)\right\}\right)\right), \tag{30}$$

for $i = 1,\ldots,L$

By simply extending AC-GNNs, another kind of GNNs are proposed (i.e., ACR-GNN(Aggregate-Combine-Readout GNN)), which can capture all the $FOC_2$ classifiers:

$$\begin{aligned} x_v^{(i)} &= \mathrm{COM}^{(i)}\left(x_v^{(i-1)}, \mathrm{AGG}^{(i)}\left(\left\{x_u^{(i-1)} \middle| u\in\mathcal{N}_G(v)\right\}\right),\right. \\ &\left.\mathrm{READ}^{(i)}\left(\left\{x_u^{(i-1)} \middle| u\in G\right\}\right)\right), \text{for } i = 1,\ldots,L, \end{aligned} \tag{31}$$

where READ = READOUT. Since the global computation can be costly, it is further proposed that just one readout function together with the final layer is enough to capture each $FOC_2$ classifier instead of adding the global readout function for each layer.

### *6.1.4. From homogeneous to heterogeneous*

Different from GNN models for homogeneous networks, GNNs for heterogeneous networks concern how to aggregate vertex features of different vertex types or connected with edge of different types. Attention mechanism is widely used in the design process of GNNs for heterogeneous networks (Wang et al., 2019a; Hu et al., 2020c).

In (Schlichtkrull and KipfPeter Bloemvan den Rianne Berg, 2018), the relational graph convolution network (R-GCN) is proposed to model large-scale relation data based on the message-passing frameworks. Different weight matrices are used for different relations in each layer to aggregate and transform hidden representations from each node's neighbourhood.

In HetGNN (Zhang et al., 2019a), a feature type-specific LSTM model is used to extract features of different types for each node, followed by another vertex-type specific LSTM model which is used to aggregate extracted feature vectors from different types of neighbours. Then, the attention mechanism is used to combine representation vectors from different types of neighbours.

Heterogeneous Graph Attention Network (HAN) is proposed in (Wang et al., 2019a), where meta paths are treated as edges between the connected two nodes. Here an attention mechanism based on meta paths and nodes is used to calculate neighbourhood aggregation vector and embedding matrix.

Heterogeneous Graph Transformer is proposed in (Hu et al., 2020c). A node type-specific attention mechanism is used and weights for different meta paths are learned automatically.

Apart from embedding different types of nodes to the same latent space, it is also proposed in HetSANN (Hong et al., 2020) to assign different latent dimensions to them. During the aggregation process, transformation matrices are applied on the feature vectors of the target node's neighbours to transform them to the same latent space with the target node. Then, the attention based aggregation process is applied on the projected neighbourhood vectors.

## 7. Theoretical foundations

Understanding different models in a universal framework can cast some insights on the design process of corresponding embedding models (like the powerful spectral filters). Thus, in this section, we will review some theoretical basis and recent understandings for models discussed above, which can benefit the further development of related algorithms.

### *7.1. Underlying kernels: graph spectral filters*

We want to show that most of the models discussed above, whether in a shallow architecture or based on graph neural networks, have some connections with graph spectral filters.

For shallow embedding models, it has been shown in (Qiu et al., 2018a) that the some neural based models (e.g. DeepWalk (Bryan et al., 2014), node2vec (Grover and Leskovec, 2016), LINE (Tang et al., 2015b)) are implicitly factorizing matrices (Table 2). Furthermore, DeepWalk matrix can also be seen as filtering (Qiu et al., 2018a).

Moreover, the convolutional operation in spectral GNNs can be interpreted as a low-pass filtering operation (Wu et al., 2019; Hoang and Maehara, 2019). The understanding can also be easily extended to spatial GNNs since the spatial aggregation operation can be transferred to the spectral domain, according to (ShumanSunil et al., 2013).

Apart from those implicitly filtering models, graph filters have also been explicitly used in ProNE (Zhang et al., 2019b), GraphZoom (Deng et al., 2020) and GraphWave (Donnat et al., 2017) to refine vertex embeddings or generate vertex embeddings preserving certain kind of vertex proximities.

#### 7.1.1. Spectral filters as feature extractors

Spectral filters can be seen and used as the effective feature extractors based on their close connection with graph spatial properties.

For example, the band-pass filter $g(\lambda) = e^{-\frac{1}{2}[(\lambda-\mu)^2-1]\theta}$ is used in ProNE (Zhang et al., 2019b) to propagate vertex embeddings obtained by factorizing a sparse matrix in the first stage. The idea for "band-pass" is inspired by the Cheeger's inequality:

$$\frac{\lambda_k}{2} \le \rho_G(k) \le O(k^2)\sqrt{\lambda_k}, \tag{32}$$

where $\rho_G(k)$ is the k-way Cheeger constant, a smaller value of which means a better k-way partition. A well-known property can be concluded from Eq. (32) when setting $\lambda_k = 0$: the number of connected components in an undirected graph is equal to the number of eigenvalue zero in the graph Laplacian (Fan and Fan, 1997). Then the band-pass filter is hoped to extract the both global and local network information from raw embeddings.

In addition, heat kernel $g_s(\lambda) = e^{-\lambda s}$ is used in GraphWave (Donnat et al., 2017) to generate wavelets for each vertex (Eq. (9)), based on which vertex embeddings are calculated via empirical characteristic functions. More importantly, structural similarities can be preserved by calculating vertex embeddings in this way.

Apart from band-pass filters and heat kernels, low-pass filters are kind of more widely used filters not only in shallow embedding models (Zhang et al., 2019b), but the aggregation matrices in GNNs can be seen as low-pass matrices and the corresponding graph convolution operations can be treated as low-pass filtering operations (Wu et al., 2019; Hoang and Maehara, 2019).

For shallow embedding models, the low-pass filter $\widetilde{h}_k(\lambda) = (1-\lambda)^k$ is used to propagate the embedding matrix $\widehat{\boldsymbol{E}}_i$ to get the refined embedding matrix $\boldsymbol{E}_i$ in the embedding refinement statement of GraphZoom (Deng et al., 2020).

As for GNNs, passing vertex features through the matrix $\widetilde{\boldsymbol{D}}^{-\frac{1}{2}}\widetilde{\boldsymbol{A}}\widetilde{\boldsymbol{D}}^{-\frac{1}{2}} = \boldsymbol{I}_N - \widetilde{\mathscr{L}}$ in GCN (Kipf and Welling, 2017), where N is the number of vertices, is equal to filtering features with the filter $h(\lambda) = 1 - \lambda$ in the spectral domain (Eq. (33)), where $\Lambda$, $\boldsymbol{U}$ are the eigenvalue matrix and eigenvector matrix of $\widetilde{\mathscr{L}}$ respectively.

$$\boldsymbol{Z} = \widetilde{\boldsymbol{D}}^{-\frac{1}{2}}\widetilde{\boldsymbol{A}}\widetilde{\boldsymbol{D}}^{-\frac{1}{2}}\boldsymbol{X\Theta} = \boldsymbol{U}(\boldsymbol{I}_N - \boldsymbol{\Lambda})\boldsymbol{U}^T\boldsymbol{X\Theta} \tag{33}$$

The filter $h(\lambda) = 1 - \lambda$ is a low-pass filter since the eigenvalues of the normalized graph Laplacian satisfy the following property:

$$0 = \lambda_0 < \lambda_1 \le \ldots \le \lambda_{max} \le 2, \tag{34}$$

and $\lambda_{max} = 2$ if and only if the graph has a bipartite subgraph as its connected component (ShumanSunil et al., 2013; Hoang and Maehara, 2019; Qiu et al., 2018a).

The low-pass property of propagation matrices in GNNs is further explored in (Hoang and Maehara, 2019; Wu et al., 2019). It is proved that two techniques can enhance the low-pass property for the filters. Let $\lambda_i(\sigma)$ be the i-th smallest generalized eigenvalue of the augmented normalized graph Laplacian $(\widetilde{\boldsymbol{D}}, \widetilde{\mathscr{L}}) = \boldsymbol{D} + \sigma\boldsymbol{I}$, then $\lambda_i(\sigma)$ is a non-negative number, and monotonically decreases as the non-negative value σ increases ($\lambda_i(\sigma) = 0$ for all the $\sigma > 0$ if $\lambda_i(0) = 0$). Thus, the high frequency components will be gradually attenuated by the corresponding spectral filtering operation as σ increases. Besides, increasing the power of the graph filter $h_k(\lambda) = (1-\lambda)^k$ (i.e., stacking several GCN layers or directly using the filter $h_k(\lambda) = (1-\lambda)^k, k > 1$ in SGC (Wu et al., 2019) and gfNN (Hoang and Maehara, 2019)) can help increase the low-pass property of the spectral filter (Fig. 7 Left channel).

Moreover, the propagation matrix used in gfNN is k-th power of the augmented random walk adjacency matrix $\widetilde{\boldsymbol{A}}_{rw}^k = (\widetilde{\boldsymbol{D}}^{-1}\widetilde{\boldsymbol{A}})^k$, whose corresponding spectral filter is $h_k(\lambda) = (1-\lambda)^k$, where λ is the generalized eigenvalues and matrix $\boldsymbol{\Lambda} = \boldsymbol{U}^T\widetilde{\boldsymbol{D}}\widetilde{\boldsymbol{L}}_{rw}\boldsymbol{U}$. $\boldsymbol{U}$ is the generalized eigenvector matrix with the property $\boldsymbol{U}^T\widetilde{\boldsymbol{D}}\boldsymbol{U} = \boldsymbol{I}$. The generalized eigenpair $(\lambda, u)$ satisfies $\boldsymbol{L}u = \lambda\widetilde{\boldsymbol{D}}u$. They are also solutions of the generalized eigenvalue problem in variation form (Hoang and Maehara, 2019), which aims to find $u_1, \ldots, u_n \in \mathbb{R}^n$ such that for each $i \in 1, \ldots, n$, $u_i$ is a solution of the following optimization problem:

$$\text{minmize} \;\; \Delta(u) \;\; \text{subject to} \;\; (u, u)_{\widetilde{D}} = 1, \, (u, u_j)_{\widetilde{D}} = 0, \\ j \in 1, \ldots, n \tag{35}$$

where $\Delta(u) = u^T\boldsymbol{L}u$ is the variation of the signal u and $(u, u_j)_{\widetilde{D}} = u^T\widetilde{\boldsymbol{D}}u$ is the inner product between signal u and $u_j$. If $(\lambda, u)$ is a generalized eigenpair, then $(\lambda, \widetilde{\boldsymbol{D}}^{1/2}u)$ is an eigenpair of $\widetilde{\mathscr{L}}$.

Compared with the eigenvalues and eigenvectors of the normalized graph Laplacian, generalized eigenvectors with smaller generalized eigenvalues are smoother in terms of the variation Δ.

**Solution for Optimization Problems.** The embedding refinement problem in GraphZoom has seen that the low-pass filter matrix can serve as the close form solution of the optimization problem related with Laplacian regularization.

Another example is the Label Propagation (LP) problem for graph based semi-supervised learning (BengioOlivier and Le Roux, 2006; Zhu et al., 2003; Zhou et al., 2004), the close form of whose optimization objective function (Eq. (36)) is Eq. (37), where Y is the label matrix and Z is the objective of LP that is consistent with the label matrix Y as well as being smoothed on the graph to force nearby vertices to have similar embeddings.

$$\boldsymbol{Z} = \text{argmin}_{\boldsymbol{Z}}\|\boldsymbol{Z} - \boldsymbol{Y}\|_2^2 + \alpha\cdot\text{tr}(\boldsymbol{Z}^T\boldsymbol{L}\boldsymbol{Z}) \tag{36}$$

$$\boldsymbol{Z} = (\boldsymbol{I} + \alpha\boldsymbol{L})^{-1}\boldsymbol{Y} \tag{37}$$

#### 7.1.2. Spectral filters as kernels of matrices being factorized

We want to show that some matrices being factorized by matrix factorization algorithms can also be seen as filter matrices.

Take the matrix factorized by DeepWalk (Bryan et al., 2014) as an example. It has been shown in (Qiu et al., 2018a) that the matrix term $\left(\frac{1}{T}\sum_{r=1}^{T}\boldsymbol{P}^r\right)\boldsymbol{D}^{-1}$ in DeepWalk's matrix can be written as

$$\left(\frac{1}{T}\sum_{r=1}^{T}\boldsymbol{P}^r\right)\boldsymbol{D}^{-1} = \left(\boldsymbol{D}^{-\frac{1}{2}}\right)\left(\boldsymbol{U}\left(\frac{1}{T}\sum_{r=1}^{T}\boldsymbol{\Lambda}^r\right)\boldsymbol{U}^T\right)\left(\boldsymbol{D}^{-\frac{1}{2}}\right), \tag{38}$$

where $\boldsymbol{\Lambda}$, $\boldsymbol{U}$ are the eigenvalue matrix and eigenvector matrix of the matrix $\boldsymbol{D}^{-\frac{1}{2}}\boldsymbol{A}\boldsymbol{D}^{-\frac{1}{2}} = \boldsymbol{I} - \mathscr{L}$ respectively. The matrix $\boldsymbol{U}\left(\frac{1}{T}\sum_{r=1}^{T}\boldsymbol{\Lambda}^r\right)\boldsymbol{U}^T$ has eigenvalues $\frac{1}{T}\sum_{r=1}^{T}\lambda_i^r, i = 1, \ldots, n$, where $\lambda_i, i = 1, \ldots, n$ are eigenvalues of the matrix $\boldsymbol{D}^{-\frac{1}{2}}\boldsymbol{A}\boldsymbol{D}^{-\frac{1}{2}}$, which can be seen as the transformation (a kind of filter) applied on the eigenvalue $\lambda_i$. This filter has the two properties (Fig. 8): (1) it prefers positive large eigenvalues; (2) the preference becomes stronger as the T (the window size) increases.

Besides, we also want to show the relationship between the DeepWalk matrix and the its corresponding normalized graph Laplacian. Since $\boldsymbol{D}^{-1}\boldsymbol{A} = \boldsymbol{I} - \boldsymbol{L}_{rw}$ and the eigenvectors and eigenvalues of $\boldsymbol{L}_{rw}$ and $\mathscr{L}$ have the following relationship: if $(\lambda_i, u_i)$ is an eigenvalue-eigenvector pair for normalized graph Laplacian $\mathscr{L}$, then $L_{rw}D^{-\frac{1}{2}}u_i = \lambda_i D^{-\frac{1}{2}}u_i$. Thus, we can write $L_{rw}$ in the following form (Eq. (39)) for $\mathscr{L}$ can be written as $\mathscr{L} = U\Lambda U^T$.

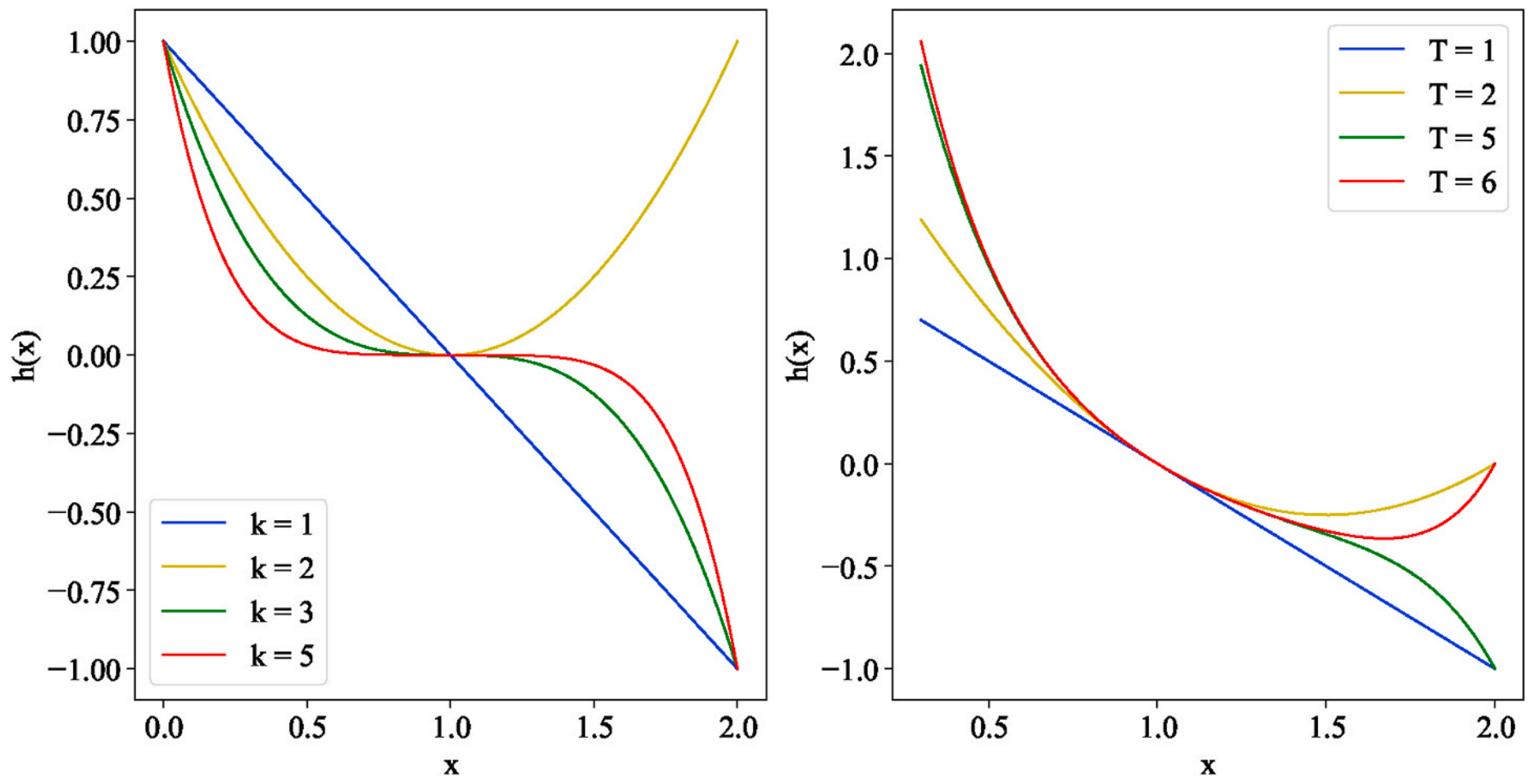

**Fig. 7.** Image of the filter function $h_k(x) = (1-x)^k$, $k \in \{1,2,3,5\}$ (Left Panel) and DeepWalk matrix filter function $h(x) = -(-x^{-1} + 1 + x^{-1}(1-x)^{T+1})$, $T \in \{1,2,5,6\}$ (Right Panel). Left Panel: Increasing the value of k can increase the band-stop characteristic of the filter function. Right Panel: The effect of increasing the window size T on the filter function.

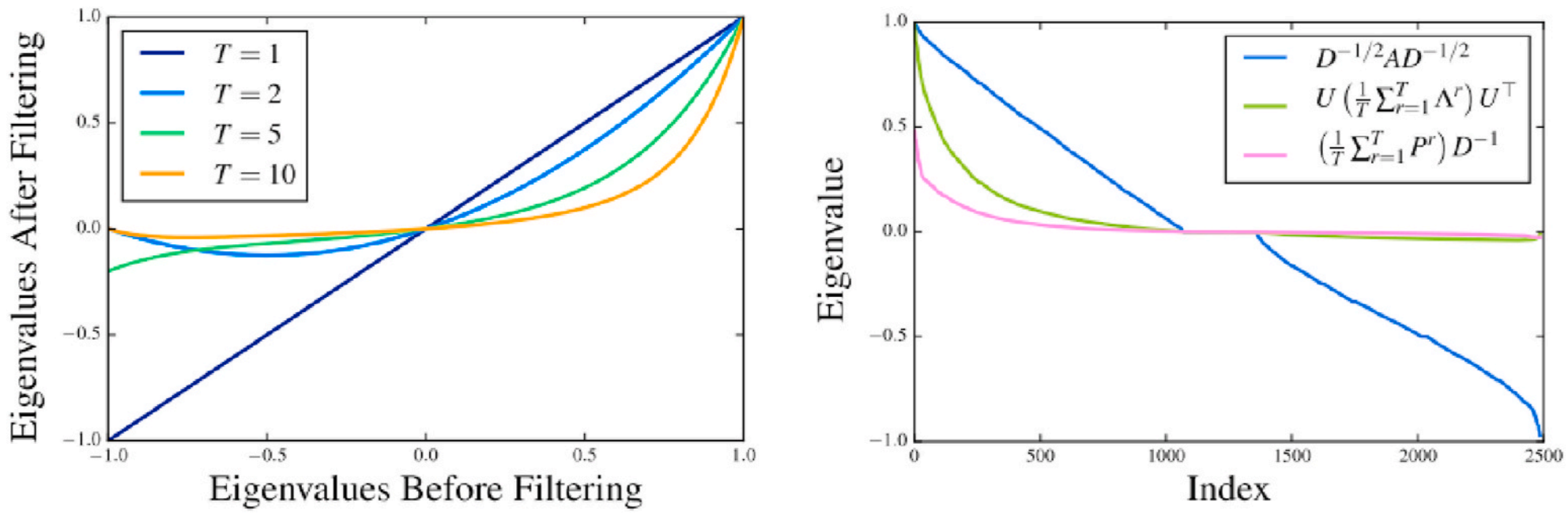

**Fig. 8.** Filtering characteristic of DeepWalk matrix's function. Left Panel: Image of the function $f(x) = \frac{1}{T}\sum_{r=1}^{T} x^r$ with **dom**$f = [-1,1]$, $T = 1,2,5,10$. Right Panel: Eigenvalues of $D^{-\frac{1}{2}}AD^{-\frac{1}{2}}$, $U\left(\frac{1}{T}\sum_{r=1}^{T}\Lambda^r\right)U^T$, $\left(\frac{1}{T}\sum_{r=1}^{T}P^r\right)D^{-1}$ for Cora network ($T = 10$). Reprinted from (Qiu et al., 2018a).

$$L_{rw} = D^{-\frac{1}{2}}U\Lambda U^T D^{\frac{1}{2}} \tag{39}$$

For each r the term $(D^{-1}A)^r$ can be written as $(I - L_{rw})^r = I - C_r^1 L_{rw} + C_r^2 L_{rw}^2 + \ldots + (-1)^r L_{rw}^r$, adding all the terms for different rs up and using a simple property for combinational number: $C_{n-1}^{m-1} + C_{n-1}^{m} = C_n^m$, we can have:

$$\begin{aligned}\sum_{r=1}^{T}(D^{-1}A)^r &= TI - C_{T+1}^2 L_{rw} + C_{T+1}^3 L_{rw}^2 \\ &+ \ldots + (-1)^T C_{T+1}^{T+1} L_{rw}^T.\end{aligned} \tag{40}$$

Then viewing Eq. (40) as the binomial expansion with some transformation, it is equal to:

$$-D^{-\frac{1}{2}}U\left(-\Lambda^{-1} + I + \Lambda^{-1}(I-\Lambda)^{T+1}\right)U^T D^{\frac{1}{2}} \tag{41}$$

Then the equivalent filter function in the spectral domain can be written as $h(\lambda) = -(-\lambda^{-1} + 1 + \lambda^{-1}(1-\lambda)^{T+1})$, which can be seen as a low-pass filter (Fig. 7, Right channel).

The matrix in the log term of DeepWalk's matrix (Table 2 DeepWalk) can be written as $-D^{-\frac{1}{2}}U(-\Lambda^{-1} + I + \Lambda^{-1}(I-\Lambda)^{T+1})U^T D^{-\frac{1}{2}}$, where $U$ is the eigenvector matrix for the normalized graph Laplacian.

The matrix factorized by LINE (Tang et al., 2015b) is a trivial example. Ignoring constant items and taking the matrix in the log item, we can have the following form $D^{-1}AD^{-1}$, which is equal to $D^{-\frac{1}{2}}(I - \mathscr{L})D^{-\frac{1}{2}}$. $(I - \mathscr{L})$ can be seen as the filter matrix with the filter function $h(\lambda) = 1 - \lambda$, and the multiplication items $D^{-\frac{1}{2}}$ can be seen as the signal rescaling matrices.

### 7.2. Universal attention mechanism

Attention mechanisms are both explicitly and implicitly widely used in many algorithms.

For shallow embedding models, the positive sampling strategy, like sliding a window in the sampled node sequences obtained by different kind of random walks on the graph (Bryan et al., 2014; Grover and Leskovec, 2016) or just sample the adjacent nodes for each target node (Tang et al., 2015b), can be seen as applying different attention weights on different nodes. Meanwhile, the negative sampling distribution given a positive sampling distribution for each node, which can also be seen as performing attention mechanism on different nodes, is proposed with the negative sampling strategy in (Mikolov et al., 2013b) and further discussed in (Yang et al., 2020b). It is also closely related with the performance of the generated embedding vectors.

The positive samples and negative samples are then used in the optimization process and can help maintain the corresponding node proximities.

For GNNs, following the idea in an unsubmitted manuscript (Anonymous, 2020), by writing the aggregation function in the following universal form:

$$\boldsymbol{H} = \sigma(\mathscr{N}(\mathscr{L}\boldsymbol{Q}\boldsymbol{H}\boldsymbol{W}), \tag{42}$$

where Q is a diagonal matrix, $\mathscr{L}$ is the matrix related with graph adjacency matrix, $\mathscr{N}(\cdot)$ is the normalization function, $\sigma(\cdot)$ is the nonlinearity transformation function perhaps with post-propagation rescaling, the aggregation process of graph neural networks can be interpreted and separated as the following four stages: pre-propagation signal rescaling, propagate, re-normalization, and post-propagation signal rescaling.

The proposed paradigm for GNNs can help with the design of neural architecture search algorithms for GNNs. The search space can be designed based on the pre- and post-propagation rescaling matrices, the normalization function $\mathscr{N}(\cdot)$, and feature transformation function $\sigma(\cdot)$, and so on.

The pre-propagation signal rescaling process is usually used as the attention mechanism in many GCN variants, like the attention mechanism in GAT (Veličković et al., 2018), whose pre-propagation rescaling scheme can be seen as multiplying a diagonal matrix $\boldsymbol{Q}$ to the right side of the matrix $\mathscr{L} = \boldsymbol{A} + \boldsymbol{I}_N$, with each of the element in matrix $Q_{jj} = (\boldsymbol{W}\boldsymbol{H}\overrightarrow{q})_j$, where $\overrightarrow{q} \in \mathbb{R}^{d\times 1}$ is a trainable parameter. Besides, the post-propagation signal rescaling can also be combined with attention mechanism, which can be seen as left multiplying a diagonal matrix $\boldsymbol{P}$ to the propagated signal matrix $\mathscr{N}(\mathscr{L}\boldsymbol{Q})\boldsymbol{H}\boldsymbol{W}$. For example, the edge attention can be performed by setting $\boldsymbol{P} = \frac{1}{\mathscr{L}Q\overrightarrow{1}}$, k-hop edge attention can be performed in the similar form: $\boldsymbol{P} = \frac{1}{\mathscr{L}^k Q\overrightarrow{1}}$, matrix $\boldsymbol{P}$ for k-hop path attention can have the following form:

$$\mathscr{L}' = \mathscr{L}\boldsymbol{Q}_k\mathscr{L}\boldsymbol{Q}_{k-1}\ldots\mathscr{L}\boldsymbol{Q}_1, \overrightarrow{p} = \frac{1}{\mathscr{L}'\overrightarrow{1}}. \tag{43}$$

Moreover, the aggregation process in RGCN (Zhu et al., 2019), where features with large variance can be attenuated, can also be seen as the attention process applied on vertex feature variance and can help improve the robustness of GCN.

## 8. Optimization methods

The optimization strategies we choose for a certain embedding model will affect its time and space efficiency and even the quality of embedding vectors. Some tricks and rethinkings can help reinforce the theoretical basis of related algorithms and further improve their expressiveness as well as reduce the time consumption. Thus, in this section, we will review optimization strategies of some typical embedding models, which are also of great importance in the design process.

### 8.1. Optimizations for random walk based models

Optimization strategies for random walk based models can start with those for natural language embedding problems, which focus on word sequences. We refer these optimization problems to Sequence Optimization Problems, which can date bask to the classical N-gram models. But since the calculation complexity will increase significantly as the word sequence's length grows (Bryan et al., 2014), the problem is relaxed in (Mikolov et al., 2013a) with two proposed models, CBOW and Skip-Gram.

**Skip-Gram.** We will focus on the Skip-Gram model, which uses the target node to predict context nodes by maximizing the following probability:

$$\Pr(\{v_{i-w}, \ldots, v_{i+w}\} / \{v_i\} | \Phi(v_i)), \tag{44}$$

where $\{v_{i-w}, \ldots, v_{i+w}\}$ are context vertices chosen by a sliding window over the node sequence with window size w. By applying the i.i.d. assumption and ignoring the order of context nodes, Eq. (44) can be factorized to

$$\prod_{i-w\le j\le i+w, j\ne i} \Pr\left(v_j \middle| \Phi(v_i)\right). \tag{45}$$

The order independent assumption can better capture the "nearness" in graph structures (Bryan et al., 2014).

However, the calculation for the probability $\Pr(v_j|\Phi(v_i))$ is also time-consuming ($O(|N|)$, N is the number of nodes in the graph), which is always in the form of softmax (Eq. (46)). Thus, two strategies are proposed to alleviate this problem, that is hierarchical softmax (Mikolov et al., 2013a) and negative sampling (Mikolov et al., 2013b).

#### 8.1.1. Hierarchical softmax

In hierarchical softmax (Fig. 9), node embeddings that need optimizing are organized into a binary tree and then calculation for Eq. (46) is transferred into the multiplication of softmax in the path from root to the target context vertex (Eq. (47)). Then the time complexity can be reduced to $O(\log|V|)$.

$$\Pr\left(v_j \middle| \Phi(v_i)\right) = \frac{e^{\alpha_i\cdot\beta_j}}{\sum_{v_k\in\mathscr{V}} e^{\alpha_i\cdot\beta_k}} \tag{46}$$

$$\Pr\left(u_k|\Phi(v_i)\right) = \prod_{l=1}^{\log V} \Pr\left(b_l \middle| \Phi\left(v_j\right)\right) \tag{47}$$

#### 8.1.2. Negative sampling

An alternative for hierarchical softmax is Noise Contrastive Estimation (NCE) as proposed in (Gutmann and Hyvärinen, 2012), based on which the Negative Sampling strategy is introduced in (Mikolov et al., 2013b) and has gain a wide application (Bryan et al., 2014; Grover and Leskovec, 2016; Tang et al., 2015b).

The objective of NCE can be shown to approximately maximize the log probability of the softmax (Mikolov et al., 2013b). Negative Sampling strategy is a simplified NCE concerned only with learning high-quality vector representations. Eq. (48), where $w_I$ is the target word, $w_O$ is the context word, $w_i$ is the sampled negative word, $v'_{w_O}$ is the context embedding vector for word $w_O$ and $v_{w_I}$ is the word embedding vector for word $w_I$, $P_n(\cdot)$ is the negative sampling distribution, can be used to replace every $\log P(w_O|w_I)$ term in the Skip-Gram objective. Maximizing Eq. (45) can be seen as distinguishing the target word $w_O$ from k negative samples drew from the noise distribution $P_n(w)$ using logistic regression.

$$\log\sigma\left(v'^{T}_{w_O} v_{w_I}\right) + \sum_{i=1}^{k} \mathbb{E}_{w_i\sim P_n(w)}\left[\log\sigma\left(-v'^{T}_{w_i} v_{w_I}\right)\right] \tag{48}$$

The negative sampling distribution $P_n(w)$ is a free parameter, which is set to the unigram distribution $U(w)$ raised to 3/4 rd power (i.e., $U(w)^{3/4}/Z$) in (Mikolov et al., 2013b) since it can outperform significantly the unigram and the uniform distributions.

**Further Understanding for Negative Sampling (NS).** However, neither the unigram distribution nor the 3/4 rd power of the unigram distribution that is employed in word2vec (Mikolov et al., 2013b) is the best negative sampling distribution. In (Yang et al., 2020b), it is theoretically proved that the best negative sampling distribution should be positively but sub-linearly correlated to the corresponding positive sampling distribution.

Although the proposed design principle seems contrary to the intuition that nodes with high positive sampling rates should have lower negative sampling rates, NS distribution designed by following this theory can indeed improve the performance of existing algorithms.

In (Levy and Goldberg, 2014), it is proved that the optimal dot-product of the word-context pair's representations should take the form of PMI (Eq. (4)) by replacing the NS distribution with the uniform distribution. Keeping the negative sampling distribution term, the close form of $\overrightarrow{u}^T\overrightarrow{v}$, where $\overrightarrow{v}$ is the representation vector of node v, is as follows:

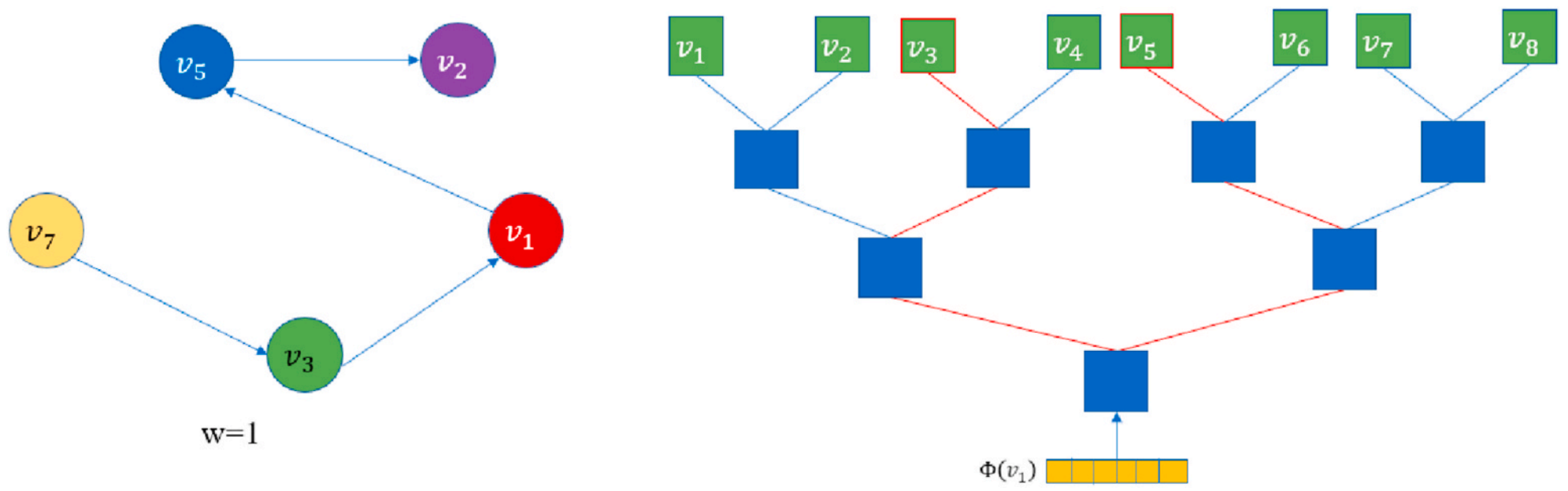

**Fig. 9.** An illustration for hierarchical softmax.Left Panel: A toy example of random walk with window size $w = 1$. Right Panel: Suppose we want to maximize the probability $\Pr(v_3|\Phi(v_1))$ and $\Pr(v_5|\Phi(v_1))$, they are factorized out over sequences of probability distributions corresponding to the paths starting at the root and ending at $v_3$ and $v_5$ respectively. Adapted from (Bryan et al., 2014).

$$\overrightarrow{u}^T\overrightarrow{v} = -\log\left(\frac{k\cdot p_n(u|v)}{p_d(u|v)}\right), \tag{49}$$

from which it can be seen that the positive and negative sampling distribution ($p_d$ and $p_n$) have the same level influence on embedding results.

The effect of negative sampling can be further seen from the process of derivating the optimal NS distribution. The best NS distribution should minimize the gap between the theoretical solution of corresponding parameters: $\boldsymbol{\theta} = [\overrightarrow{u}_0^T\overrightarrow{v}, \ldots, \overrightarrow{u}_{N-1}^T\overrightarrow{v}]$, and their empirical solution, since only limited positive and negative samples can be obtained in practice. One way is to minimize the mean square error between the theoretical and empirical results of $\overrightarrow{u}^T\overrightarrow{v}$:

$$\mathbb{E}\left[||(\boldsymbol{\theta}_T - \boldsymbol{\theta}^*)_u||^2\right] = \frac{1}{T}\left(\frac{1}{p_d(u|v)} - 1 + \frac{1}{kp_n(u|v)} - \frac{1}{k}\right). \tag{50}$$

It claims that nodes with high positive sampling rates should also be negatively sampled sufficiently, otherwise the expectation of the mean square error would increase.

By setting the NS rate for node u given node v positively but sublinearly correlated to its positive sampling rate, i.e., $p_n(u|v) \propto p_d^\alpha(u|v)$, the dot-product $\overrightarrow{u}^T\overrightarrow{v}$ can also have the following monotonicity:

$$\overrightarrow{u}_i^T\overrightarrow{v} = \log p_d(u_i|v) - \alpha \log p_d(u_i|v) + c \tag{51}$$

$$> (1-\alpha)\log p_d(u_j|v) + c = \overrightarrow{u}_j^T\overrightarrow{v},$$

if $p_d(u_i|v) > p_d(u_j|v)$.

### 8.2. Optimization strategies for matrix factorization

#### 8.2.1. SVDs

SVD is a basic algorithm from linear algebra which is used to factorize matrix $\boldsymbol{M}$ into the product of three matrices $\boldsymbol{U}\cdot\boldsymbol{\Sigma}\cdot\boldsymbol{V}^T$, where $\boldsymbol{\Sigma}$ is the diagonal singular value matrix, $\boldsymbol{U}$ and $\boldsymbol{V}$ are orthogonal matrices with each of their row vector as an unit vector. The time complexity of the economic SVD on $m \times n$ matrix is $O(mn\min(m,n))$.

One property is that the optimal rank d approximation of matrix $\boldsymbol{M}$ can be obtained by taking $\boldsymbol{M}_d = \boldsymbol{U}_d\cdot\boldsymbol{\Sigma}_d\cdot\boldsymbol{V}_d^T$, where $\boldsymbol{\Sigma}_d$ is the diagonal matrix formed from the top d singular values, $\boldsymbol{U}_d$ and $\boldsymbol{V}_d$ are matrices formed by selecting the corresponding columns from $\boldsymbol{U}$ and $\boldsymbol{V}$. To be specific, $\boldsymbol{M}_d = \mathrm{argmin}_{\mathrm{Rank}(\boldsymbol{M}')=d}\boldsymbol{M}' - \boldsymbol{M}_2$.

Truncated Singular Value Decomposition (tSVD) aims to find the matrix factorization results $\boldsymbol{U}_d, \boldsymbol{\Sigma}_d, \boldsymbol{V}_d$ given a specific d.

**r-tSVD.** Randomized matrix method can be used to accelerate the basic tSVD algorithm. The idea is to find an orthogonal matrix $\boldsymbol{Q}$ by performing the iterative QR decomposition on the random projected matrix $\boldsymbol{H} = \boldsymbol{M\Omega}$, where $\boldsymbol{\Omega}$ is a Gaussian random matrix with fewer columns than $\boldsymbol{M}$. Then tSVD can be preformed on the matrix $\boldsymbol{B} = \boldsymbol{Q}^T\boldsymbol{M}$, which will cost less time than performing tSVD directly on matrix $\boldsymbol{M}$, since $\boldsymbol{B}$ is a small matrix with less rows than $\boldsymbol{M}$. Then $\boldsymbol{U}_d = \boldsymbol{QU}_d, \boldsymbol{\Sigma}_d, \boldsymbol{V}_d$ are the approximated results of tSVD on the matrix $\boldsymbol{M}$, where $[\boldsymbol{U}_d, \boldsymbol{\Sigma}_d, \boldsymbol{V}_d] = \mathrm{tSVD}(\boldsymbol{B})$. r-tSVD is used in the sparse matrix factorization stage in ProNE (Zhang et al., 2019b), whose time complexity is linear with respect to the number of edges in the graph (i.e., $O(|E|)$).

#### 8.2.2. Acceleration of rPCA for sparse matrix

PCA is similar with tSVD when being used to factorize a specific matrix $\boldsymbol{M}$ (Xu et al., 2018). Some methods can be used to accelerate the calculation process of the basic rPCA algorithm due to the sparse matrices' properties, like replacing the QR decomposition in the iteration process of rPCA with LU decomposition, replacing the Gaussian random matrix Ω with the uniform random matrix, and so on.

For example, two acceleration versions (frPCA and its variant frPCAt) of rPCA are proposed in (Xu et al., 2018) based on the modified power iteration scheme, which is faster than rPCA and can also provide more flexible trade-off between runtime and accuracy. Actually, they can accelerate ProNE by about 5–10 times.

#### 8.2.3. Iterative updating strategy

Apart from tSVD and its variants, the target matrix can be obtained by iteratively updating corresponding matrices after the optimization objective is defined, like algorithmes for non-negative matrix factorization proposed in (Lee and Seung, 2001) and the joint non-negative matrix factorization method stated in (Akata et al., 2011). Their applications can be seen in M-NMF (Wang et al., 2017b).

### 8.3. Optimization strategies for GNNs

Optimization strategies for GNNs focus on defining corresponding objective functions, which are then optimized under the stochastic gradient descent paradigm. The objective function can be supervised or unsupervised, which can be defined according to downstream tasks. In this subsection, we will only focus on the design of unsupervised objective functions.

**Vertex Proximity Based.** For example, the Laplacian regularization term (Eq. (52)) is widely used in (Zhu et al., 2003; Zhou et al., 2004; Belkin et al., 2006) based on the assumption that connected nodes tend to share the same label.

$$O_{\mathrm{reg}} = \sum_{i,j} A_{ij}\|f(X_i) - f(X_j)\|^2 = f(\boldsymbol{X})^T\boldsymbol{L}f(\boldsymbol{X}) \tag{52}$$

However, as stated in (Kipf and Welling, 2017), this assumption may restrict the modeling capacity, since graph edges are always half observed and are not necessarily encode node similarity.

The random walk-like loss function is used in (Hamilton et al., 2017) to encourage vertices with high co-occurrence frequency to have similar representations:

$$O(z_u) = -\log\left(\sigma\left(z_u^T z_v\right)\right) - Q \cdot \mathbb{E}_{v_n \sim P_n(v)} \log\left(\sigma\left(-z_u^T z_{v_n}\right)\right), \tag{53}$$

where $z_u$ is output representation of node u, v is a node that co-occurs near u in a fixed-length random walk.

**Others.** Graph information can also be utilized differently. For instance, Graph Infomax (Velickovic et al., 2019) and InfoGraph (Sun et al., 2019) maximize the mutual information between vertex representations and the pooled graph representation. In Variational Graph Auto-Encoder (Kipf and Welling, 2016), vertex representations are used to reconstruct the graph structure.

## 9. Challenges and problems

### *9.1. Shallow embedding modles*

**Random Walk Based Models.** Despite the theoretical bound lying behind shallow embedding models, which give them connections with the matrix factorization models (Qiu et al., 2018a). The equivalent can be satisfied only when the walk length goes to infinite, which leading that fact that random walk based models cannot outperform matrix factorization based methods, which has also been shown empirically (Qiu et al., 2018a). Moreover, the sampling process is time-consuming if high order proximities are wished to be preserved (Rozemberczki and Sarkar, 2020).

**Matrix Factorization Based Models.** Factorizing matrices which encode high-order node proximities, structural information and other side information is guaranteed to obtain high-quality vertex embeddings. However, factorizing large, dense matrices is still time-consuming, though it has been proved that the factorizing process can be accelerated by random matrix theory when the matrix is sparse (Zhang et al., 2019b; Xu et al., 2018). And matrices being factorized are destined to be dense ones if high-order vertex proximities, and structural information are wished to be preserved.

**Summary.** Ways to solve problems of shallow embedding models mentioned above can be found in an embedding model ProNE (Zhang et al., 2019b), where a spare matrix is designed to be factorized to get raw node embeddings, then the spectral propagation process is applied on the obtained embeddings to make sure that the propagated embeddings are aware of high-order vertex and structural information. Since the matrix being factorized is a spare one, and the propagation process is also economical based on some mathematical properties, the model ProNE is a fast and effective model, combining advantages of different embedding models and also staying time-efficient (Fig. 10).

However, there are also some inherent problems lying in shallow embedding models:

In the first place, the look-up embedding table in shallow neural embedding models and matrices in matrix factorization based embedding methods decide that those models are inherently transductive. Generating embedding vectors for new nodes needs to be calculated from scratch or it will take a long time. Moreover, if there are encoders in such models, they are relatively simple, and it is hard to incorporate vertex content in the encoding process. Even though the deep neural encoders are adopted in DNGR (Cao et al., 2016) and SDNE (Wang et al., 2016), features that are fed into the encoders are $|V|$ - dimensional connectivity proximity vectors and the reconstruction architecture makes it hard to encode vertex content with connectivity information.

Those problems can be partly solved by Graph Neural Network based models.

### *9.2. Graph neural networks*

Compared with shallow embedding models, GNNs present their potential in exploring graph structures (Xu et al., 2019b), exploiting content information of nodes and edges (Veličković et al., 2018; Vaswani et al., 2017), being inductive (Cen et al., 2019), as well as dealing with heterogeneous structures better (Cen et al., 2019; Zhang et al., 2019a; Wang et al., 2019a). Although they can solve problems of shallow embedding models to some extend, generating node features that are more meaningful, better aware of node structural information and content information, there are some inherent problems lying in GNNs' architecture (Zhao and Akoglu, 2019; Yu et al., 2019b; Zhu et al., 2019; Deng et al., 2019): (1) GNN models always tend to increase the number of GNN layers to capture information from high-order neighbours, which can lead to three problems:

| Dataset | training ratio | 0.1 | 0.3 | 0.5 | 0.7 | 0.9 |
|---|---|---|---|---|---|---|
| PPI | DeepWalk | 16.4 | 19.4 | 21.1 | 22.3 | 22.7 |
| | LINE | 16.3 | 20.1 | 21.5 | 22.7 | 23.1 |
| | node2vec | 16.2 | 19.7 | 21.6 | 23.1 | 24.1 |
| | GraRep | 15.4 | 18.9 | 20.2 | 20.4 | 20.9 |
| | HOPE | 16.4 | 19.8 | 21.0 | 21.7 | 22.5 |
| | ProNE (SMF) | 15.8 | 20.6 | 22.7 | 23.7 | 24.2 |
| | ProNE | **18.2** | **22.7** | **24.6** | **25.4** | **25.9** |
| | (±σ) | (±0.5) | (±0.3) | (±0.7) | (±1.0) | (±1.1) |
| Wiki | DeepWalk | 40.4 | 45.9 | 48.5 | 49.1 | 49.4 |
| | LINE | **47.8** | 50.4 | 51.2 | 51.6 | 52.4 |
| | node2vec | 45.6 | 47.0 | 48.2 | 49.6 | 50.0 |
| | GraRep | 47.2 | 49.7 | 50.6 | 50.9 | 51.8 |
| | HOPE | 38.5 | 39.8 | 40.1 | 40.1 | 40.1 |
| | ProNE (SMF) | 47.6 | 51.6 | 53.2 | 53.5 | 53.9 |
| | ProNE | 47.3 | **53.1** | **54.7** | **55.2** | **57.2** |
| | (±σ) | (±0.7) | (±0.4) | (±0.8) | (±0.8) | (±1.3) |
| BlogCatalog | DeepWalk | 36.2 | 39.6 | 40.9 | 41.4 | 42.2 |
| | LINE | 28.2 | 30.6 | 33.2 | 35.5 | 36.8 |
| | node2vec | **36.3** | 39.7 | 41.1 | 42.0 | 42.1 |
| | GraRep | 34.0 | 32.5 | 33.3 | 33.7 | 34.1 |
| | HOPE | 30.7 | 33.4 | 34.3 | 35.0 | 35.3 |
| | ProNE (SMF) | 34.6 | 37.6 | 38.6 | 39.3 | 39.0 |
| | ProNE | 36.2 | **40.0** | **41.2** | **42.1** | **42.7** |
| | (±σ) | (±0.5) | (±0.3) | (±0.6) | (±0.7) | (±1.2) |
| Dataset | training ratio | 0.01 | 0.03 | 0.05 | 0.07 | 0.09 |
| DBLP | DeepWalk | 49.3 | 55.0 | 57.1 | 57.9 | 58.4 |
| | LINE | 48.7 | 52.6 | 53.5 | 54.1 | 54.5 |
| | node2vec | 48.9 | 55.1 | 57.0 | 58.0 | 58.4 |
| | GraRep | 50.5 | 52.6 | 53.2 | 53.5 | 53.8 |
| | HOPE | **52.2** | 55.0 | 55.9 | 56.3 | 56.6 |
| | ProNE (SMF) | 50.8 | 54.9 | 56.1 | 56.7 | 57.0 |
| | ProNE | 48.8 | **56.2** | **58.0** | **58.8** | **59.2** |
| | (±σ) | (±1.0) | (±0.5) | (±0.2) | (±0.2) | (±0.1) |
| Youtube | DeepWalk | 38.0 | 40.1 | 41.3 | 42.1 | 42.8 |
| | LINE | 33.2 | 35.5 | 37.0 | 38.2 | 39.3 |
| | ProNE (SMF) | 36.5 | 40.2 | 41.2 | 41.7 | 42.1 |
| | ProNE | **38.2** | **41.4** | **42.3** | **42.9** | **43.3** |
| | (±σ) | (±0.8) | (±0.3) | (±0.2) | (±0.2) | (±0.2) |

**Fig. 10.** The classification performance in terms of Micro-F1 (%). Reprint from (Zhang et al., 2019b).

- over-fitting. Since the signal propagation process is always coupled with non-linear transformation. Thus, increasing the number of layers will lead to increasing the number of parameters at the same time, which will raise the risk of over-fitting;
- over-smoothing, it has been proved that the convolution operation is essentially a special form of Laplacian smoothing (Li et al., 2018). The spatial form of the convolution operation centered at a target node is just linearly aggregating features from its neighbourhood. Thus, directly stacking many layers will make each node incorporate too many features from others but lose specific information of itself (Chen et al., 2019; Zhao and Akoglu, 2019), leading to the over-smoothing problem;
- non-robust, which comes from the propensity to over-fitting towards noisy part of input features (Hoang and Maehara, 2019) as the number of parameters increases.

(2) The propagation process in GNN models will always make each node too dependent on its neighbours, thus leading to the non-robust problem as stated above. Moreover, since edges in real-world networks are always only partially observed, even with false edges, it will make the model be more sensitive to adversarial attack on graph data and hard to learn true features for each node (Zhu et al., 2019). The attacker can indirectly attack the target node by just manipulating long-distance neighbours (Zügner et al., 2018). (3) Moreover, different from shallow embedding models, like random walk based models, matrix factorization based models, which rely more on graph structures the advantages that GNNs have over shallow embedding like easily incorporating with node and edge features, labels, also make GNNs rely on labels too much. Thus it is hard for GNNs to perform well when there are only scarce labels available. (4) Designing the best GNN for a certain task requires manual tuning to adjust the network architecture and hyper-parameters, such as the attention mechanism in the neighbourhood aggregation process, the activation functions and the number of hidden dimensions. Apart from those problems regard to the architecture of GNN models mentioned above, huge parameters in GNNs that require tuning also lead to the heavy manual labors in GNNs' design process, which is also a common problem in deep learning community.

**Proposed Solutions.** Problems stated above are also opportunities to help us design better models. There are many works aiming to tackle the problems for graph neural networks (Zhao and Akoglu, 2019; Yu et al., 2019b; Zhu et al., 2019; Deng et al., 2019; Verma et al., 2019). We will discuss some basic insights later.

#### *9.2.1. Graph regularization*

**Random Propagation.** Just as the effectiveness of Dropout for the training process of neural networks, which can be seen an adaptive $L_2$ regularization strategy, introducing random factors in the feature propagation process of GNNs (Yu et al., 2019b; Feng et al., 2020) has been proved as an effective way to help improve the model's robustness, alleviate over-fitting and over-smoothing problems. In DropEdge (Yu et al., 2019b), a certain rate of edges will be droped out in each training epoch, while in DropNode (Feng et al., 2020), a certain rate of nodes' features will be droped out. Both of them can help block the some information propagation ways, reduce vertices' dependency on certain neighbours and thus can help the model go deep, alleviate the over-smoothing problem and improve the robustness.

Compared with DropEdge, dropping out the entire features of some nodes in the training epoch can further help decouple the feature propagation and transformation process, which are closely connected in the deterministic GNN architectures. Moreover, the consistency loss is introduced to help force the model output similar predictions with different augmentations (with different nodes randomly dropped) as input, whose effectiveness is empirically proved.

**Data Augmentation.** Data augmentation can help increase data varieties, thus helping avoid over-fitting problems.

For example, a graph data interpolation method is used in GraphMix (Verma et al., 2019) to augment data. Moreover, the random propagation (Yu et al., 2019b; Feng et al., 2020) can also be seen as a data augmentation technique since random deformed copies of the original graph can increase the randomness and diversity of the input data. Thus, it can help increase data varieties and help avoid over-fitting problems.

Moreover, several data augmentation techniques are proposed in (Wang et al., 2020) to improve the performance of GNNs during inference time. A consistency loss is introduced and minimized to make the prediction of node labels under different augmentations consistent with each other:

$$\mathscr{L}_C = \frac{1}{|\mathscr{V}|}\sum_{i\in\mathscr{V}}\mathscr{D}_{KL}\left(p\left(y\middle|\mathscr{G}_i, x_j j\in\mathscr{V}_i, \widetilde{\Theta}\right) \parallel p(y|\widehat{\mathscr{G}}_i, \widehat{x}_j j\in\widehat{\mathscr{V}}_i, \widetilde{\Theta}\right)\Big), \quad (54)$$

where $\mathscr{G}_i$ is the subgraph of node $v_i$, which corresponds to the receptive field of node $v_i$, $\mathscr{V}_i$ is the node set of the subgraph, $x_i$ denotes the attributes of node $v_i$, $\Theta$ is the parameter set. $\widehat{\mathscr{G}}_i$ is the subgraph of node $v_i$ after augmentation, $\widetilde{\Theta}$ is a fixed copy of the current parameter $\Theta$, indicating that the gradient is not propagated through $\Theta$.

In addition, a "parallel universe" data augmentation scheme is introduced to conduct data augmentation for different nodes individually and separately, since the modification on the subgraph of a node $v_i$ will influence the input features of other nodes.

**Adversarial Virtual attack.** Performing adversarial virtual attacks and introducing corresponding loss term to the optimization objective can help improve the smoothness of output vectors with respect to perturbation around the local structure (Deng et al., 2019).

BVAT. In (Deng et al., 2019), two virtual attack strategies are proposed to improve the adversarial virtual attack techniques in VAT (Miyato et al., 2018), to get close to worst-case attack for each node. The first one is S-BVAT, where a set of nodes $\mathscr{V}_S \subset \mathscr{V}$ is sampled. The receptive field for each node in the set will not overlap with other nodes' receptive fields. Then the regularization term for training is the average LDS loss over nodes in $\mathscr{V}_S$:

$$\mathscr{R}_{vadv}(\mathscr{V}_S, \mathscr{W}) = \frac{1}{B}\sum_{u\in\mathscr{V}_S}\text{LDS}(\boldsymbol{X}_u, \mathscr{W}, r_{vadv,u}), \quad (55)$$

where $\mathscr{W}$ is the trainable parameters, $\boldsymbol{X}_u$ is the input feature matrix of all nodes in node u's receptive field. Then, $\mathscr{R}_{vadv}(\mathscr{V}_{\mathscr{S}}, \mathscr{W})$ can be seen as an approximate estimation of $\mathscr{R}_{vadv}(\mathscr{V}, \mathscr{W})$. The second one is called O-BVAT, where the average LDS loss with respect to the whole perturbation matrix $\boldsymbol{R}$ corresponding to the whole feature matrix $\boldsymbol{X}$ is maximized by solving:

$$\max_{\boldsymbol{R}}\frac{1}{N}\sum_{u\in\mathscr{V}}D_{KL}\left(p\left(y\middle|\boldsymbol{X}_u, \widehat{\mathscr{W}}\right) \parallel p(y|\boldsymbol{X}_u + \boldsymbol{R}_u, \mathscr{W}\right)\Big) - \gamma\cdot \parallel \boldsymbol{R}\parallel_F^2, \quad (56)$$

where $\|\boldsymbol{R}\|_F$ is the Frobenius norm of $\boldsymbol{R}$ to make the optimal perturbation have a small norm and $\gamma$ is a hyper-parameter to balance the loss terms. $\boldsymbol{R}$ is optimized for T iterations, which is more powerful than one-step gradient-based methods.

RGCN. Different from VAT and BVAT, which introduce perturbation to the training process deliberately, a robust GCN model (RGCN) is introduced to improve model's robustness by reducing aggregation of features with large variance. The basic idea of RGCN is replacing direct message propagation processes in GCN and message passing GNN models with the convolutional operation on Gaussian distributions. The technique to avoid attacking on GNNs models was an attention mechanism, which assigns different weights to features according their variances since larger variances may indicate more uncertainties in the latent representations and larger probability of having been attacked (Zhu et al., 2019). Meanwhile, the "reparameterization trick" (Doersch, 2016) is used to optimize the loss function using back propagation and an explicit regularization term was used to constrain the latent representations in the first layer to Gaussian distribution:

$$\mathscr{L}_{reg1} = \sum_{i=1}^{N} KL(\mathscr{N}(\boldsymbol{\mu}_i^{(1)}, \text{diag}(\boldsymbol{\sigma}_i^{(1)})) \parallel \mathscr{N}(0, \boldsymbol{I})), \quad (57)$$

where $\boldsymbol{\mu}_i^{(1)}$ is the mean vector for i-th feature in the first layer, $\boldsymbol{\sigma}_i^{(1)}$ is the variance vector for i-th feature in the first layer, $\mathscr{N}(\mu, \sigma)$ is the normal distribution with $\mu$ as its mean vector and $\sigma$ as its covariance matrix, and $KL(\cdot \parallel \cdot)$ is the KL-divergence between two distributions.

**Summary.** Fig. 11 taken from (Feng et al., 2020) summarizes and compares the experiment results with regard to different GNN models'

| Category | Method | Cora | Citeseer | Pubmed |
|---|---|---|---|---|
| Graph Convolution | GCN [27] | 81.5 | 70.3 | 79.0 |
| | GAT [42] | 83.0±0.7 | 72.5±0.7 | 79.0 ±0.3 |
| | Graph U-Net [15] | 84.4±0.6 | 73.2±0.5 | 79.6±0.2 |
| | MixHop [2] | 81.9 ± 0.4 | 71.4±0.8 | 80.8±0.6 |
| | GMNN [36] | 83.7 | 72.9 | 81.8 |
| | GraphNAS [16] | 84.2±1.0 | 73.1±0.9 | 79.6±0.4 |
| Regularization based GCNs[2] | VBAT [13] | 83.6 ± 0.5 | 74.0 ± 0.6 | 79.9 ± 0.4 |
| | $G^3$NN [31] | 82.5 ±0.2 | 74.4±0.3 | 77.9 ± 0.4 |
| | GraphMix [43] | 83.9 ± 0.6 | 74.5 ±0.6 | 81.0 ± 0.6 |
| | DropEdge [37] | 82.8 | 72.3 | 79.6 |
| Sampling based GCNs[3] | GraphSAGE [22] | 78.9±0.8 | 67.4±0.7 | 77.8±0.6 |
| | FastGCN [9] | 81.4±0.5 | 68.8±0.9 | 77.6±0.5 |
| Our methods | GRAND | **85.4±0.4** | **75.4±0.4** | **82.7±0.6** |
| | GRAND_GCN | 84.5±0.3 | 74.2±0.3 | 80.0±0.3 |
| | GRAND_GAT | 84.3±0.4 | 73.2± 0.4 | 79.2±0.6 |
| | GRAND_dropout | 84.9±0.4 | 75.0±0.3 | 81.7±1.0 |

**Fig. 11.** Experiment results of several GNN models in node classification tasks on three public datasets. "Our methods" denotes GRAND model (Feng et al., 2020). Reprint from (Feng et al., 2020).

effectiveness on several public datasets in node classification task. Moreover, it is also shown that graph regularization techniques can address non-robust, over-smoothing and over-fitting problems to some extend, which demonstrates the effectiveness of these methods as a way to improve GNN models.

### 9.2.2. *Self-supervised learning for GNNs*

Self-supervised learning (SSL) is an effective technique that is widely adopted in NLP and computer vision domain to extract expressive representations for words, sentences and images.

Recently, SSL on graph data has attracted a lot of interests, especially for GNNs.

Self-supervised learning for GNNs focuses on defining proper pretext tasks and training techniques.

Basic pretext tasks are defined on characteristics of graph data, such as attribute completion (Hu et al., 2020a; JinTyler et al., 2020; Zhang et al., 2020; Zhu et al., 2020), edge prediction (Hu et al., 2020a; JinTyler et al., 2020; Zhu et al., 2020). Some models try to define pretext tasks based on graph topology, like vertex distance prediction (JinTyler et al., 2020), context prediction (Hu et al., 2020a), graph structure recovery (Zhang et al., 2020), pair-wise proximity prediction (Peng et al., 2020), and so on. Moreover, Graph Contrastive Coding (GCC) is proposed in (Qiu et al., 2020) based on the contrastive learning paradigm (He et al., 2019; Chen et al., 2020). A generative pre-training model (GPT-GNN) is proposed in (Hu et al., 2020b) to pre-train GNNs based on vertex attributes generation and edge generation tasks. Labels are also combined in the pretext tasks in (JinTyler et al., 2020) to align with down stream tasks.

As for training techniques, pre-training under pretext tasks and fine-tuning on down stream tasks is a widely used paradigm (Hu et al., 2020a; Zhang et al., 2020). Besides, self-training is also a effective training technique and is used in (You et al., 2020). It pre-trains a model in the labeled data and then uses it to generate pseudo-labels for unlabeled data, which are included into labeled data for the next round of training (Zoph et al., 2020). Moreover, the multi-task training can also be defined to combine self-supervised pretext tasks and down stream supervised tasks.

In conclusion, self-supervised learning broadens the idea of training GNNs and expands our exploration space.

### 9.2.3. *Neural architecture search for GNNs*

Aiming to alleviate the heavy manual tuning labors lying in GNNs' design process, the developing of neural architecture search (NAS) on GNN models focus on how to automatically design GNN architectures (Zhou et al., 2019). Designing the best GNN for a certain task requires manual tuning to adjust the network architecture and hyper-parameters, such as the attention mechanism in the neighbourhood aggregation process, the activation functions and the number of hidden dimensions. Thus, the developing of neural architecture search (NAS) on GNN models focus on how to automatically design GNN architectures (Zhou et al., 2019).

Although neural architecture search has rose a lot of interests (Elsken et al., 2019) and has outpreformed handcrafted ones at many other domains or tasks (e.g., image classification (Zoph and Quoc Le, 2017; Zoph et al., 2018), image generation (Wang and Huan, 2019)), it is not a trivial thing to generate such strategies to design auto-GNN models as stated in (Zhou et al., 2019).

A neural architecture search model should define the searching space, a controller which is used to judge the most prosperous models and are also supposed to do efficient parameter sharing, which can avoid training a new model from scratch. However, as for the search space, search space for GNNs is different from those of existing NAS work. For example, the search space in CNN models involves the kernel size and the number of convolutional layers, while in GNN models, the search space is defined on the activation functions, aggregation strategies, etc. Moreover, the traditional controller is inefficient to discover the potentially well-performed GNN architectures due to the inherent property of GNN models which determines the performance of GNNs varies significantly with slight architecture modification. The problem partly comes from the difficulty of evaluating GNN models, which reveals the importance of understanding GNN models deeply and defining good evaluation methods as a recent paper does (Ribeiro TulioWu and CarlosSingh, 2020) in NLP domain. Thirdly, the widely adopted techniques such as the parameter sharing is not suitable for heterogeneous GNN models which will have different weight shapes or output statistics (Guo et al., 2019).

Thus, in (Zhou et al., 2019), an efficient controller is designed based on the property of GNNs and define the concept of heterogeneous graph neural networks and permit parameter sharing only between two homogeneous GNNs. Given the best architecture at the time, the architecture modification is realized by the following three steps: (1) For each class, remove it and treat the remaining GNN architecture as the current stage. Then, use a RNN encoder to generate actions of this class for each layer. (2) Use an action guider to sample a list of classes to be modified based on the decision entropy of each class. The decision entropy of class c is defined as follows:

$$E_c \triangleq \sum_{i=1}^{n} \sum_{j=1}^{m_c} -P_{ij}\log P_{ij}, \tag{58}$$

where n is the number of layers, $m_c$ is the number of actions that can be chosen from in class c, $\overrightarrow{P}_i$ is the action probability distribution in layer i of class c. (3) Modify the GNN architecture of each class in the generated class list.

Besides, regrading the network architecture search of deep convolutional neural networks in the field of computer vision, a recent paper (Radosavovic et al., 2020) moves from the traditional design paradigm which focuses on designing individual network instances tuned to a single setting to searching design spaces (Radosavovic et al., 1882) which aims to discover general and interpretable design principles.

The discovery of the general design principles can better guide the future design of individual well-performed networks, which is also a meaningful direction that the neural architecture search for GNNs can focus on.

## 10. Future development directions

In Section 9 we summarize the existing challenges and problems in both shallow and GNN based embedding models. Although there have already been some works trying to solve those challenges, they still cannot be addressed completely and elegantly. Designing an embedding algorithm which is both effective and efficient is not an easy thing.

In this section, we will further review challenges of designing embedding algorithms on real-world networks and also some promising developing directions, hoping to be helpful for the future development.

**Dynamic.** Networks in the real world are always evolving, such as new users (new vertices) in social networks, new citations (new edges) in citation networks. Although there are some works trying to develop embedding algorithms for evolving networks, there are also many underlying challenges in such researches since the corresponding embedding algorithms should deal with the changing networks and be able to update embedding vectors efficiently (Zhang et al., 2018).

**Robustness.** In the past two years, attacks and defenses on graph data have attracted widespread attention (Sun et al., 2020). It is shown that whether unsupervised models or models with supervision from downstream tasks can be fooled even by unnoticeable perturbations (Zügner et al., 2018; Bojchevski and Günnemann, 2019). Moreover, edges and vertices in real-world networks are always uncertain and noisy (Zhang et al., 2018). It is crucial to learn representations that are robust with respect to those uncertainties and possible adversarial attacks on graphs. Some universal techniques are widely adopted to improve the embedding robustness, like using the adversarial attack as a regularizer (e.g., ANE (Dai et al., 2017), ATGA (Pan et al., 2018) VBAT (Deng et al., 2019)), modeling graph structure using probability distribution methods (e.g., URGE (Hu et al., 2017) and RGCN (Zhu et al., 2019)).

**Generating Real-World Networks.** Generating real-world networks is a meaningful thing. For example, generating molecular graphs can help with the drug design and discovery process (Shi et al., 2020; You and LiuRex Ying, 2018), generating real-world citation networks or social networks can help design more reasonable benchmarks and defend adversarial attacks. However, designing efficient density estimation and generating models on graphs is a challenging thing due to graphs' inherent combinational property and worth researching on.

**Reasoning Ability of GNNs.** Recently, there are also some works digging into the reasoning ability of GNNs. They try to explore GNNs' potential in executing algorithms (Xu et al., 2020; Velikovi et al., 2020), or focus on the logical expressiveness of GNNs (Zhang et al., 2019c; BarcelEgor et al., 2020). Both of them can help us better understand the internal mechanism of GNNs and thus help promote the development of GNN models to generate more expressive and powerful vertex embeddings. There are also attempts trying to combine GNNs with the statistical relational learning (SRL) problem, which have been well explored in Markov networks (Taskar et al., 2012), Markov logic networks (Richardson and Domingos, 2006) usually based on conditional random field, to help GNNs model relational data (Qu et al., 2019).

## Acknowledgments

The work is supported by the National Key R&D Program of China (2018YFB1402600), NSFC for Distinguished Young Scholar (61825602), NSFC (61672313), NSFC (61836013), and Tsinghua-Bosch Joint ML Center.

## References

Aggarwal, Charu C., 2011. An introduction to social network data analytics. In: Social Network Data Analytics.
Ahmed, Nesreen K., Ryan, Rossi, John, Boaz Lee, Willke, Theodore L., Zhou, Rong, Kong, Xiangnan, Eldardiry, Hoda, 2018. Learning Role-Based Graph Embeddings.
Akata, Zeynep, Thurau, Christian, Bauckhage, Christian, 2011. Non-negative Matrix Factorization in Multimodality Data for Segmentation and Label Prediction.
Alon, Noga, Chen, Avin, Koucky, Michal, Kozma, Gady, Lotker, Zvi, Mark, R., 2007. Tuttle. Many Random Walks Are Faster than One.
Anonymous, 2020. Understanding Graph Convolutional Networks as Signal Rescaling.
B Tenenbaum, Joshua, De Silva, Vin, Langford, John C., 2000. A global geometric framework for nonlinear dimensionality reduction. Science 290 (5500), 2319–2323. https://doi.org/10.1126/science.290.5500.2319.
Barcel, Pablo, Egor, V. Kostylev, Monet, Mikael, Prez, Jorge, Reutter, Juan, Pablo Silva, Juan, 2020. The Logical Expressiveness of Graph Neural Networks. ICLR.
Belkin, Mikhail, Niyogi, Partha, 2002. Laplacian Eigenmaps and Spectral Techniques for Embedding and Clustering. NIPS.
Belkin, Mikhail, Niyogi, Partha, Sindhwani, Vikas, 2006. Manifold regularization: a geometric framework for learning from labeled and unlabeled examples. J. Mach. Learn. Res. 7 (85), 2399–2434. http://jmlr.org/papers/v7/belkin06a.html.
Bengio, Yoshua, Olivier, Delalleau, Le Roux, Nicolas, 2006. Label Propagation and Quadratic Criterion.
Bhagat, Smriti, Graham, Cormode, Muthukrishnan, S., 2011. Node classification in social networks. In: Social Network Data Analytics.
Bojchevski, Aleksandar, Günnemann, Stephan, 2019. Adversarial Attacks on Node Embeddings via Graph Poisoning. ICML.
Bruna, Joan, Zaremba, Wojciech, Arthur Szlam, LeCun, Yann, 2013. Spectral Networks and Locally Connected Networks on Graphs. arXiv.
Bruno Ribeiro, F., Donald Towsley, F., 2010. Estimating and Sampling Graphs with Multidimensional Random Walks, pp. 390–403 internet measurement conference.
Bryan, Perozzi, Al-Rfou, Rami, Skiena, Steven, 2014. Deepwalk: Online Learning of Social Representations. KDD.
Bryan, Perozzi, Kulkarni, Vivek, Chen, Haochen, Skiena, Steven, 2016. Don't Walk, Skip! Online Learning of Multi-Scale Network Embeddings.
Cai, HongYun, Zheng, Wenchen Vincent, Kevin Chang, Chen-Chuan, 2018. A Comprehensive Survey of Graph Embedding: Problems, Techniques and Applications. TKDE.
Cao, Shaosheng, Lu, Wei, Xu, Qiongkai, 2015. Grarep: Learning Graph Representations with Global Structural Information. CIKM.
Cao, Shaosheng, Lu, Wei, Xu, Qiongkai, 2016. Deep Neural Networks for Learning Graph Representations. AAAI.
Cen, Yukuo, Zou, Xu, Zhang, Jianwei, Yang, Hongxia, Zhou, Jingren, Tang, Jie, 2019. Representation Learning for Attributed Multiplex Heterogeneous Network. KDD.
Chen, Ting, Sun, Yizhou, 2017. Task-guided and Path-Augmented Heterogeneous Network Embedding for Author Identification. WSDM.
Chen, Jie, Ma, Tengfei, Xiao, Cao, 2018. Fastgcn: Fast Learning with Graph Convolutional Networks via Importance Sampling.
Chen, Deli, Lin, Yankai, Li, Wei, Peng, Li, Zhou, Jie, Sun, Xu, 2019. Measuring and Relieving the Over-smoothing Problem for Graph Neural Networks from the Topological View.
Chen, Ting, Kornblith, Simon, Norouzi, Mohammad, Hinton, Geoffrey, 2020. A Simple Framework for Contrastive Learning of Visual Representations. ICML.
Cheng, Yang, Liu, Zhiyuan, Zhao, Deli, Sun, Maosong, Chang, Edward, 2015. Network Representation Learning with Rich Text Information. IJCAI.
Church, Kenneth, Hanks, Patrick, 1990. Word association norms, mutual information, and lexicography. Computational linguistics.
Cormen, Th, Leiserson, Ce, Rivest, Rl, Stein, C., 2001. Introduction to Algorithms, second ed.
Dai, Quanyu, Li, Qiang, Tang, Jian, Wang, Dan, 2017. Adversarial Network Embedding.
David Shuman, I., Ricaud, Benjamin, Vandergheynst, Pierre, 2013. Vertex-frequency Analysis on Graphs. Applied and Computational Harmonic Analysis.
Defferrard, Michaël, Bresson, Xavier, Vandergheynst, Pierre, 2016. Convolutional Neural Networks on Graphs with Fast Localized Spectral Filtering. NIPS.
Deng, Zhijie, Dong, Yinpeng, Zhu, Jun, 2019. Batch Virtual Adversarial Training for Graph Convolutional Networks. arXiv.
Deng, Chenhui, Zhao, Zhiqiang, Wang, Yongyu, Zhang, Zhiru, Feng, Zhuo, 2020. Graphzoom: A Multi-Level Spectral Approach for Accurate and Scalable Graph Embedding. ICLR.
Doersch, Carl, 2016. Tutorial on Variational Autoencoders. arXiv.
Dong, Yuxiao, Chawla, V. Nitesh, Swami, Ananthram, 2017. metapath2vec: Scalable Representation Learning for Heterogeneous Networks. KDD.
Dong, Yuxiao, Hu, Ziniu, Wang, Kuansan, Sun, Yizhou, Tang, Jie, 2020. Heterogeneous Network Representation Learning. IJCAI.
Donnat, Claire, Zitnik, Marinka, Hallac, David, Leskovec, Jure, 2017. Spectral Graph Wavelets for Structural Role Similarity in Networks. CoRR.
Elsken, Thomas, Jan Metzen, Hendrik, Hutter, Frank, 2019. Neural architecture search: a survey. J. Mach. Learn. Res. 1–21.
Fan, RK Chung, Fan, Chung Graham, 1997. Spectral Graph Theory. American Mathematical Soc.
Feng, Wenzheng, Zhang, Jie, Dong, Yuxiao, Han, Yu, Luan, Huanbo, Xu, Qian, Yang, Qiang, Kharlamov, Evgeny, Tang, Jie, 2020. Graph Random Neural Network. NIPS.
Fragkiskos, D Malliaros, Vazirgiannis, Michalis, 2013. Clustering and community detection in directed networks: a survey. Phys. Rep. https://dblp.org/db/journals/corr/corr1308.html#MalliarosV13.
Fu, Tao-Yang, Lee, Wang-Chien, Lei, Zhen, 2017. Hin2vec: Explore Meta-Paths in Heterogeneous Information Networks for Representation Learning. CIKM.
Gao, Sheng, Denoyer, Ludovic, Gallinari, Patrick, 2011. Temporal link prediction by integrating content and structure information. In: Proceedings of the 20th ACM International Conference on Information and Knowledge Management.
Gilmer, Justin, Schoenholz, Samuel S., Riley, Patrick F., Oriol Vinyals, Dahl, George E., 2017. Neural Message Passing for Quantum Chemistry. ICML.

Gonzalez, Joseph E., Xin, Reynold S., Dave, Ankur, Crankshaw, Daniel, Franklin, Michael J., Ion Stoica, 2014. Graphx: graph processing in a distributed dataflow framework. In: 11th {USENIX} Symposium on Operating Systems Design and Implementation. {OSDI} 14.
Goodfellow, Ian J., Pouget-Abadie, Jean, Mirza, Mehdi, Xu, Bing, Warde-Farley, David, Ozair, Sherjil, Courville, Aaron, Bengio, Yoshua, 2014. Generative Adversarial Networks.
Grover, Aditya, Leskovec, Jure, 2016. node2vec: Scalable Feature Learning for Networks. KDD.
Guo, Zichao, Zhang, Xiangyu, Mu, Haoyuan, Wen, Heng, Liu, Zechun, Wei, Yichen, Sun, Jian, 2019. Single Path One-Shot Neural Architecture Search with Uniform Sampling. CVPR.
Gutmann, Michael U., Hyvärinen, Aapo, 2012. Noise-contrastive estimation of unnormalized statistical models, with applications to natural image statistics. J. Mach. Learn. Res. 13.
Hamilton, Will, Ying, Zhitao, Leskovec, Jure, 2017. Inductive representation learning on large graphs. NIPS (News Physiol. Sci.) 1024–1034.
Hammond, David K., Vandergheynst, Pierre, Gribonval, Rémi, 2011. Wavelets on Graphs via Spectral Graph Theory. Applied and Computational Harmonic Analysis.
Harris, Zellig S., 1954. Distributional structure. Word 10 (2–3), 146–162. https://www.tandfonline.com/toc/rwrd20/10/2-3.
He, Kaiming, Fan, Haoqi, Wu, Yuxin, Xie, Saining, Girshick, Ross, 2019. Momentum Contrast for Unsupervised Visual Representation Learning. CVPR.
Henderson, Keith, Gallagher, Brian, Eliassi-Rad, Tina, Tong, Hanghang, Basu, Sugato, Akoglu, Leman, Koutra, Danai, Faloutsos, Christos, Li, Lei, 2012. Rolx: Structural Role Extraction & Mining in Large Graphs. KDD.
Hoang, N.T., Maehara, Takanori, 2019. Revisiting Graph Neural Networks: All We Have Is Low-Pass Filters. arXiv.
Hochstenbach, M.E., 2009. A Jacobi-Davidson Type Method for the Generalized Singular Value Problem. Linear Algebra & Its Applications.
Hong, Huiting, Guo, Hantao, Lin, Yucheng, Yang, Xiaoqing, Zang, Li, Ye, Jieping, 2020. An Attention-Based Graph Neural Network for Heterogeneous Structural Learning. AAAI.
Hu, Pili, Lau, Cheong Wing, 2013. A Survey and Taxonomy of Graph Sampling. CoRR.
Hu, Jiafeng, Cheng, Reynold, Huang, Zhipeng, Fang, Yixang, Luo, Siqiang, 2017. On Embedding Uncertain Graphs. CIKM.
Hu, Weihua, Liu, Bowen, Gomes, Joseph, Zitnik, Marinka, Liang, Percy, Pande, Vijay, Leskovec, Jure, 2020a. Strategies for Pre-training Graph Neural Networks. ICLR.
Hu, Ziniu, Dong, Yuxiao, Wang, Kuansan, Chang, Kai-Wei, Sun, Yizhou, 2020b. Gpt-gnn: Generative Pre-training of Graph Neural Networks. KDD.
Hu, Ziniu, Dong, Yuxiao, Wang, Kuansan, Sun, Yizhou, 2020c. Heterogeneous Graph Transformer. WWW.
Huang, Wenbing, Zhang, Tong, Yu, Rong, Huang, Junzhou, 2018. Adaptive Sampling towards Fast Graph Representation Learning.
Hu, Binbin, Yuan, Fang, Shi, Chuan, 2019. Adversarial Learning on Heterogeneous Information Networks.
Ido, Dagan, Pereira, Fernando, Lee, Lillian, 1994. Similarity-based Estimation of Word Cooccurrence Probabilities. arXiv.
Jacob, Yann, Denoyer, Ludovic, Gallinari, Patrick, 2014. Learning Latent Representations of Nodes for Classifying in Heterogeneous Social Networks. WSDM.
Jin, Yilun, song, Guojie, Shi, Chuan, 2020. Gralsp: Graph Neural Networks with Local Structural Patterns. AAAI.
Jin, Wei, Tyler, Derr, Liu, Haochen, Wang, Yiqi, Wang, Suhang, Liu, Zitao, Tang, Jiliang, 2020. Self-supervised Learning on Graphs: Deep Insights and New Direction. arXiv.
Kipf, Thomas N., Welling, Max, 2016. Variational Graph Auto-Encoders. arXiv.
Kipf, Thomas N., Welling, Max, 2017. Semi-supervised Classification with Graph Convolutional Networks. ICLR.
Kiran, K Thekumparampil, Wang, Chong, Oh, Sewoong, Li, Li-Jia, 2018. Attention-based Graph Neural Network for Semi-supervised Learning. arXiv.
Krizhevsky, Alex, Sutskever, Ilya, Hinton, Geoffrey E., 2012. Imagenet Classification with Deep Convolutional Neural Networks. NIPS.
Lee, Daniel D., Seung, H Sebastian, 2001. Algorithms for Non-negative Matrix Factorization. NIPS.
Leskovec, Jure, Faloutsos, Christos, 2006. Sampling from Large Graphs. KDD, pp. 631–636.
Levy, Omer, Goldberg, Yoav, 2014. Neural Word Embedding as Implicit Matrix Factorization. NIPS.
Li, Juzheng, Zhu, Jun, Zhang, Bo, 2016. Discriminative Deep Random Walk for Network Classification. ACL.
Li, Qimai, Han, Zhichao, Wu, Xiao-Ming, 2018. Deeper Insights into Graph Convolutional Networks for Semi-supervised Learning. AAAI.
Liben-Nowell, David, Kleinberg, Jon, 2003. The link prediction problem for social networks. In: Proceedings of the Twelfth International Conference on Information and Knowledge Management.
Lin, Yankai, Liu, Zhiyuan, Sun, Maosong, Liu, Yang, Zhu, Xuan, 2015. Learning Entity and Relation Embeddings for Knowledge Graph Completion. AAAI.
Linyuan, Lü, Zhou, Tao, 2011. Link prediction in complex networks: a survey. Phys. Stat. Mech. Appl. 390 (6), 1150–1170. https://www.sciencedirect.com/science/journal/03784371/390/6.
Liu, Zemin, Zheng, Vincent W., Zhou, Zhao, Zhu, Fanwei, Kevin, Chen-Chuan Chang, Wu, Minghui, Jing, Ying, 2018. Distance-aware Dag Embedding for Proximity Search on Heterogeneous Graphs. AAAI.
Lorrain, Franois, White, Harrison C., 1977. Structural Equivalence of Individuals in Social Networks. Social Networks.
Lovász, László, et al., 1993. Random Walks on Graphs: A Survey. Combinatorics. Paul erdos is eighty.
LowY, BicksonD., et al., 2012. Distributedgraphlab: Aframeworkformachinelearninganddata Mininginthecloud. ProceedingsoftheVLDBEndowment.
Marinka, Zitnik, Rok, Sosi, Marcus, W., Feldman, Jure, Leskovec, 2019. Evolution of resilience in protein interactomes across the tree of life. Proc. Natl. Acad. Sci. U. S. A 116 (10), 4426–4433.
Martins, Ines Filipa, Teixeira, Ana L., Pinheiro, Luis, Falcao, Andre O., 2012. A bayesian approach to in silico blood-brain barrier penetration modeling. J. Chem. Inf. Model. 52 (6), 1686–1697.
Micali, Silvio, Allen, Zeyuan Zhu, 2016. Reconstructing markov processes from independent and anonymous experiments. Discrete Appl. Math. 200, 108–122. https://www.sciencedirect.com/science/journal/0166218X/200/supp/C.
Mikolov, Tomas, Chen, Kai, Corrado, Greg, Dean, Jeffrey, 2013a. Efficient Estimation of Word Representations in Vector Space.
Mikolov, Tomas, Sutskever, Ilya, Chen, Kai, Corrado, Greg S., Dean, Jeff, 2013b. Distributed Representations of Words and Phrases and Their Compositionality. NIPS.
Miyato, Takeru, Maeda, Shin-ichi, Koyama, Masanori, Shin, Ishii, 2018. Virtual Adversarial Training: a Regularization Method for Supervised and Semi-supervised Learning. PAMI.
Myers, Seth A., Sharma, Aneesh, Gupta, Pankaj, Lin, Jimmy, 2014. Information Network or Social Network? the Structure of the Twitter Follow Graph. WWW.
Natarajan, Nagarajan, Inderjit, S Dhillon, 2014. Inductive matrix completion for predicting gene–disease associations. Bioinformatics 30 (12), i60–i68. https://www.researchgate.net/journal/Bioinformatics-1460-2059.
Ou, Mingdong, Cui, Peng, Pei, Jian, Zhang, Ziwei, Zhu, Wenwu, 2016. Asymmetric Transitivity Preserving Graph Embedding. KDD.
Page, L., 1998. The pagerank citation ranking : bringing order to the web. online manuscript. http://www-db.stanford.edu/backrub/pageranksub.ps.
Pan, Shirui, Hu, Ruiqi, Long, Guodong, Jiang, Jing, Yao, Lina, Zhang, Chengqi, 2018. Adversarially Regularized Graph Autoencoder for Graph Embedding.
Peng, Zhen, Dong, Yixiang, Luo, Minnan, Wu, Xiao-Ming, Zheng, Qinghua, 2020. Self-supervised Graph Representation Learning via Global Context Prediction. arXiv.
Pizarro, Narciso, 2007. Structural identity and equivalence of individuals in social networks: beyond duality. Int. Sociol. 22 (6), 767–792.
Qiu, Jiezhong, Dong, Yuxiao, Ma, Hao, Li, Jian, Wang, Kuansan, Tang, Jie, 2018a. Network Embedding as Matrix Factorization: Unifying Deepwalk, Line, Pte, and Node2vec. WSDM.
Qiu, Jiezhong, Dong, Yuxiao, Ma, Hao, Li, Jian, Wang, Chi, Wang, Kuansan, Tang, Jie, 2019. Netsmf: Large-Scale Network Embedding as Sparse Matrix Factorization. WWW.
Qiu, Jiezhong, Chen, Qibin, Dong, Yuxiao, Zhang, Jing, Yang, Hongxia, Ding, Ming, Wang, Kuansan, Tang, Jie, 2020. GCC: Graph Contrastive Coding for Graph Neural Network Pre-training. KDD.
Qu, Meng, Bengio, Yoshua, Tang, Jian, 2019. Gmnn: Graph Markov Neural Networks.
Radosavovic, Ilija, Johnson, Justin, Xie, Saining, Lo, Wan-Yen, Piotr Dollaacute, r, 1882–1890. On Network Design Spaces for Visual Recognition. ICCV, p. 2019.
Radosavovic, Ilija, Raj Kosaraju, Prateek, Girshick, Ross, He, Kaiming, DollÃ¡r, Piotr, 2020. Designing Network Design Spaces. CVPR.
Ribeiro, Leonardo FR., Saverese, Pedro HP., Figueiredo, Daniel R., 2017. struc2vec: learning node representations from structural identity. In: KDD.
Ribeiro Tulio, Marco, Wu, Tongshuang, Carlos, Guestrin, Singh, Sameer, 2020. Beyond Accuracy: Behavioral Testing of Nlp Models with Checklist. ACL.
Richardson, M., Domingos, P., 2006. Markov Logic Networks (Vol 62, Pg 107, 2006). Machine Learning.
Roweis, Sam T., Lawrence, K Saul, 2000. Nonlinear dimensionality reduction by locally linear embedding. Science 290 (5500), 2323–2326.
Rozemberczki, Benedek, Sarkar, Rik, 2020. Fast Sequence-Based Embedding with Diffusion Graphs.
Schlichtkrull, Michael, Kipf, N. Thomas, Peter Bloem, van den Rianne Berg, 2018. Ivan Titov, and Max Welling. Modeling Relational Data with Graph Convolutional Networks. ESWC.
Scott, C Ritchie, Watts, Stephen, Fearnley, Liam G., Holt, Kathryn E., Abraham, Gad, Inouye, Michael, 2016. A scalable permutation approach reveals replication and preservation patterns of network modules in large datasets. Cell systems 3 (1), 71–82.
Sen, Prithviraj, Namata, Galileo, Bilgic, Mustafa, Getoor, Lise, Galligher, Brian, Eliassi-Rad, Tina, 2008. Collective Classification in Network Data. AI Magazine.
Shervashidze, Nino, Schweitzer, Pascal, Jan van Leeuwen, Erik, Mehlhorn, Kurt, Karsten, M Borgwardt, 2011. Weisfeiler-lehman graph kernels. J. Mach. Learn. Res. 12.
Shi, Chuan, Hu, Binbin, Zhao, Xin Wayne, Philip Yu, S., 2019. Heterogeneous Information Network Embedding for Recommendation. TKDE.
Shi, Chence, Xu, Minkai, Zhu, Zhaocheng, Zhang, Weinan, Zhang, Ming, Tang, Jian, 2020. Graphaf: a Flow-Based Autoregressive Model for Molecular Graph Generation. ICLR.
Shuman, David I., Ricaud, Benjamin, Vandergheynst, Pierre, 2016. Vertex-frequency Analysis on Graphs. Applied and Computational Harmonic Analysis.
Shuman, David I., Sunil, K Narang, Pascal, Frossard, Ortega, Antonio, Vandergheynst, Pierre, 2013. The Emerging Field of Signal Processing on Graphs: Extending High-Dimensional Data Analysis to Networks and Other Irregular Domains. IEEE.
Sun, Fan-Yun, Hoffmann, Jordan, Tang, Jian, 2019. Infograph: Unsupervised and Semi-supervised Graph-Level Representation Learning via Mutual Information Maximization. arXiv.
Sun, Lichao, Dou, Yingtong, Yang, Carl, Wang, Ji, Philip, S Yu, Li, Bo, 2020. Adversarial Attack and Defense on Graph Data: A Survey. arXiv.

Tang, Jian, Qu, Meng, Mei, Qiaozhu, 2015a. Pte: Predictive Text Embedding through Large-Scale Heterogeneous Text Networks. KDD.
Tang, Jian, Qu, Meng, Wang, Mingzhe, Zhang, Ming, Yan, Jun, Mei, Qiaozhu, 2015b. Line: large-scale information network embedding. In: WWW.
Tang, Jian, Liu, Jingzhou, Zhang, Ming, Mei, Qiaozhu, 2016. Visualizing Large-Scale and High-Dimensional Data. WWW.
Taskar, Ben, Abbeel, Pieter, Koller, Daphne, 2012. Discriminative probabilistic models for relational data. Proc.conf.on Uncertainty in Artificial Intelligence 485–492.
Taylor, Graham W., Fergus, Rob, LeCun, Yann, Bregler, Christoph, 2010. Convolutional learning of spatio-temporal features. In: European Conference on Computer Vision.
Tremblay, Nicolas, Borgnat, Pierre, 2014. Graph Wavelets for Multiscale Community Mining. IEEE Transactions on Signal Processing.
Turney, Peter D., 2001. Mining the web for synonyms: pmi-ir versus lsa on toefl. In: European Conference on Machine Learning.
Turney, Peter D., Pantel, Patrick, 2010. From frequency to meaning: vector space models of semantics. J. Artif. Intell. Res. 37 (1). https://www.researchgate.net/journal/Journal-of-Artificial-Intelligence-Research-1076-9757.
Vaswani, Ashish, Shazeer, Noam, Parmar, Niki, Uszkoreit, Jakob, Jones, Llion, Gomez, Aidan N., Kaiser, Łukasz, Polosukhin, Illia, 2017. Attention Is All You Need. NIPS.
Veličković, Petar, Cucurull, Guillem, Casanova, Arantxa, Romero, Adriana, Lio, Pietro, Bengio, Yoshua, 2018. Graph Attention Networks. ICLR.
Velickovic, Petar, Fedus, William, Hamilton, William L., Pietro, Liò, Bengio, Yoshua, Hjelm, R Devon, 2019. Deep Graph Infomax. ICLR.
Velikovi, Petar, Ying, Rex, Padovano, Matilde, Hadsell, Raia, Blundell, Charles, 2020. Neural Execution of Graph Algorithms. ICLR.
Verma, Vikas, Qu, Meng, Lamb, Alex, Bengio, Yoshua, Kannala, Juho, Tang, Jian, 2019. Graphmix: Regularized Training of Graph Neural Networks for Semi-supervised Learning. arXiv.
Von Luxburg, Ulrike, 2007. A Tutorial on Spectral Clustering. Statistics and computing.
Wang, Hanchao, Huan, Jun, 2019. Agan: towards Automated Design of Generative Adversarial Networks. CoRR.
Wang, Daixin, Cui, Peng, Zhu, Wenwu, 2016. Structural Deep Network Embedding. KDD.
Wang, Hongwei, Wang, Jia, Wang, Jialin, Zhao, Miao, Zhang, Weinan, Zhang, Fuzheng, Xie, Xing, Guo, Minyi, 2017a. Graphgan: Graph Representation Learning with Generative Adversarial Nets. TKDE.
Wang, Xiao, Cui, Peng, Wang, Jing, Pei, Jian, Zhu, Wenwu, Yang, Shiqiang, 2017b. Community preserving network embedding. AAAI.
Wang, Xiao, Ji, Houye, Shi, Chuan, Wang, Bai, Cui, Peng, Yu, S Philip, Ye, Yanfang, 2019a. Heterogeneous Graph Attention Network. WWW.
Wang, Zekai, Liu, Hongzhi, Du, Yingpeng, Wu, Zhonghai, Zhang, Xing, 2019b. Unified Embedding Model over Heterogeneous Information Network for Personalized Recommendation. IJCAI, pp. 203–209.
Wang, Yiwei, Wang, Wei, Liang, Yuxuan, Cai, Yujun, Liu, Juncheng, Bryan, Hooi, 2020. Nodeaug: Semi-supervised Node Classification with Data Augmentation. KDD.
Weisfeiler, Boris, Lehman, Andrei A., 1968. A Reduction of a Graph to a Canonical Form and an Algebra Arising during This Reduction. Nauchno-Technicheskaya Informatsia.
West, D.B., 2001. Introduction to Graph Theory, second ed. Networks.
Wu, Felix, Zhang, Tianyi, Amauri Holanda de Souza Jr., Fifty, Christopher, Tao, Yu, Weinberger, Kilian Q., 2019. Simplifying Graph Convolutional Networks. ICML.
Xie, Min, Yin, Hongzhi, Wang, Hao, Xu, Fanjiang, Chen, Weitong, Wang, Sen, 2016. Learning graph-based poi embedding for location-based recommendation. In: Proceedings of the 25th ACM International on Conference on Information and Knowledge Management.
Xu, Feng, Xie, Yuyang, Song, Mingye, Yu, Wenjian, Tang, Jie, 2018. Fast Randomized Pca for Sparse Data.
Xu, Bingbing, Shen, Huawei, Qi, Cao, Qiu, Yunqi, Cheng, Xueqi, 2019a. Graph Wavelet Neural Network. ICLR.
Xu, Keyulu, Hu, Weihua, Leskovec, Jure, Jegelka, Stefanie, 2019b. How Powerful Are Graph Neural Networks? ICLR.
Xu, Keyulu, Li, Jingling, Zhang, Mozhi, Du, Simon S., Kawarabayashi, Ken-ichi, Jegelka, Stefanie, 2020. What Can Neural Networks Reason about? ICLR.
Yan, S., Xu, D., Zhang, B., Zhang, H., Yang, Q., Lin, S., 2007. Graph Embedding and Extensions: A General Framework for Dimensionality Reduction. PAMI.
Yang, Jaewon, Leskovec, Jure, 2015. Defining and Evaluating Network Communities Based on Ground-Truth. Knowledge and Information Systems.
Yang, Carl, Xiao, Yuxin, Zhang, Yu, Sun, Yizhou, Han, Jiawei, 2020a. Heterogeneous Network Representation Learning: Survey, Benchmark, Evaluation, and beyond.
Yang, Zhen, Ding, Ming, Chang, Zhou, Yang, Hongxia, Zhou, Jingren, Tang, Jie, 2020b. Understanding Negative Sampling in Graph Representation Learning.
Yoon, Kim, 2014. Convolutional Neural Networks for Sentence Classification. arXiv.
You, Jiaxuan, Liu, Bowen, Rex Ying, S., 2018. Vijay Pande, and Jure Leskovec. Graph Convolutional Policy Network for Goal-Directed Molecular Graph Generation. NeurIPS.
You, Yuning, Chen, Tianlong, Wang, Zhangyang, Shen, Yang, 2020. When Does Self-Supervision Help Graph Convolutional Networks? arXiv.
Yu, He, Song, Yangqiu, Li, Jianxin, Cheng, Ji, Peng, Jian, Peng, Hao, 2019a. Hetespaceywalk: a heterogeneous spacey random walk for heterogeneous information network embedding. In: Proceedings of the 28th ACM International Conference on Information and Knowledge Management.
Yu, Rong, Huang, Wenbing, Xu, Tingyang, Huang, Junzhou, 2019b. Dropedge: towards Deep Graph Convolutional Networks on Node Classification.
Zeng, Hanqing, Zhou, Hongkuan, Srivastava, Ajitesh, Kannan, Rajgopal, Prasanna, Viktor, 2020. Graphsaint: Graph Sampling Based Inductive Learning Method. ICLR.
Zhang, Daokun, Yin, Jie, Zhu, Xingquan, Zhang, Chengqi, 2016a. Collective classification via discriminative matrix factorization on sparsely labeled networks. In: Proceedings of the 25th ACM International on Conference on Information and Knowledge Management.
Zhang, Daokun, Yin, Jie, Zhu, Xingquan, Zhang, Chengqi, 2016b. Homophily, Structure, and Content Augmented Network Representation Learning. ICDM.
Zhang, Xia, Chen, Weizheng, Yan, Hongfei, 2016c. Tline: scalable transductive network embedding. In: Asia Information Retrieval Symposium.
Zhang, Chao, Zhang, Keyang, Quan, Yuan, Peng, Haoruo, Zheng, Yu, Tim Hanratty, Wang, Shaowen, Han, Jiawei, 2017. Regions, Periods, Activities: Uncovering Urban Dynamics via Cross-Modal Representation Learning. WWW.
Zhang, Daokun, Yin, Jie, Zhu, Xingquan, Zhang, Chengqi, 2018. Network Representation Learning: A Survey. IEEE transactions on Big Data.
Zhang, Chuxu, Song, Dongjin, Huang, Chao, Swami, Ananthram, Nitesh Chawla, V., 2019a. Heterogeneous Graph Neural Network.
Zhang, Jie, Dong, Yuxiao, Wang, Yan, Tang, Jie, Ding, Ming, 2019b. Prone: Fast and Scalable Network Representation Learning. IJCAI.
Zhang, Yuyu, Chen, Xinshi, Yuan, Yang, Ramamurthy, Arun, Li, Bo, Qi, Yuan, Song, Le, 2019c. Can Graph Neural Networks Help Logic Reasoning?.
Zhang, Jiawei, Zhang, Haopeng, Sun, Li, Xia, Congying, 2020. Graph-bert: Only Attention Is Needed for Learning Graph Representations. arXiv.
Zhao, Lingxiao, Akoglu, Leman, 2019. Pairnorm: Tackling Oversmoothing in Gnns.
Zhou, Dengyong, Bousquet, Olivier, Lal, Thomas N., Weston, Jason, Schölkopf, Bernhard, 2004. Learning with Local and Global Consistency. NIPS.
Zhou, Kaixiong, Song, Qingquan, Huang, Xiao, Hu, Xia, 2019. Auto-gnn: Neural Architecture Search of Graph Neural Networks.
Zhu, Xiaojin, Ghahramani, Zoubin, Lafferty, John D., 2003. Semi-supervised Learning Using Gaussian Fields and Harmonic Functions. ICML.
Zhu, Shenghuo, Yu, Kai, Yun, Chi, Gong, Yihong, 2007. Combining Content and Link for Classification Using Matrix Factorization. SIGIR.
Zhu, Dingyuan, Zhang, Ziwei, Cui, Peng, Zhu, Wenwu, 2019. Robust Graph Convolutional Networks against Adversarial Attacks. KDD.
Zhu, Qikui, Du, Bo, Yan, Pingkun, 2020. Self-supervised Training of Graph Convolutional Networks. arXiv.
Zoph, Barret, Quoc Le, V., 2017. Neural Architecture Search with Reinforcement Learning. ICLR.
Zoph, Barret, Vasudevan, Vijay, Shlens, Jonathon, Quoc Le, V., 2018. Learning Transferable Architectures for Scalable Image Recognition. Computer Vision and Pattern Recognition.
Zoph, Barret, Ghiasi, Golnaz, Lin, Tsung-Yi, Cui, Yin, Liu, Hanxiao, Cubuk, Ekin D., Quoc, V. Le, 2020. Rethinking Pre-training and Self-Training. CVPR.
Zügner, Daniel, Akbarnejad, Amir, Günnemann, Stephan, 2018. Adversarial Attacks on Neural Networks for Graph Data. KDD.